\pdfoutput=1

\documentclass{CUP-JNL-NLP}

\usepackage{times}
\usepackage[T1]{fontenc}
\usepackage[utf8]{inputenc}
\usepackage{microtype}
\usepackage{inconsolata}
\usepackage[
  hypertexnames=false,
  colorlinks=true,
  linkcolor=blue,
  citecolor=blue,
  urlcolor=blue
]{hyperref}
\usepackage{amsmath}
\usepackage{float}
\usepackage{array,multirow,graphicx}
\usepackage{xcolor}
\usepackage{pifont}
\usepackage{etoolbox}
\usepackage{pgfplots}
\pgfplotsset{compat=1.16}
\usepackage{enumitem}
\setlist[itemize]{
    leftmargin=*,
    label=--,
    topsep=6pt,
    partopsep=0pt,
    itemsep=3pt,
    parsep=3pt
}
\setlist[itemize,2]{
    leftmargin=2em,
    label=\textbullet
}

\usepackage{makecell}
\usepackage{booktabs}
\usepackage{siunitx}
\usepackage{longtable}

\usepackage{tikz}
\usetikzlibrary{patterns, patterns.meta}

\newcommand{\cmark}{\ding{51}} 
\newcommand{\xmark}{\ding{55}} 

\newcommand{\bert}{\textsc{BERT}}
\newcommand{\tabibert}{\textsc{T}abi\bert}
\newcommand{\tabibench}{\textsc{T}abi\textsc{B}ench}
\newcommand{\berturk}{\bert urk}
\newcommand{\ytubert}{\textsc{ytu-c}osmos-\bert}
\newcommand{\tweetbert}{\textsc{T}urkish\bert weet}
\newcommand{\mmbert}{mm\bert}
\newcommand{\modernbert}{Modern\bert}

\newcommand{\convberturk}{\textsc{C}onv\bert urk}

\newcommand{\llmjpmodernbert}{\textsc{llm-jp-modernbert}}

\newcounter{movedcontent}
\long\def\movedcontentlist{}

\newcommand{\movetoappendix}[2]{%
  \stepcounter{movedcontent}%
  \expandafter\gdef\expandafter\movedcontentlist\expandafter{%
    \movedcontentlist
    \arabic{movedcontent} & #2 & \parbox[t]{7cm}{#1} \\
  }%
  \fbox{\textit{[Moved to Appendix]: #2}}%
}

\newcommand{\displaymovedcontent}{%
  \ifx\movedcontentlist\empty
    \multicolumn{3}{|c|}{\textit{No content moved yet}} \\
  \else
    \movedcontentlist
  \fi
}

\renewcommand{\bfseries}{\fontseries{b}\selectfont}
\robustify\bfseries
\newrobustcmd{\B}{\bfseries}

\usepackage{subcaption}

\begin{document}
\label{firstpage}

\lefttitle{\tabibert: A Large-Scale \modernbert\ Foundation Model and A Unified Benchmark for Turkish}
\righttitle{Natural Language Processing}

\papertitle{\tabibert: A Large-Scale \modernbert\ Foundation Model and A Unified Benchmark for Turkish}

\jnlPage{\pageref{firstpage}}{\pageref{lastpage}}
\jnlDoiYr{2026}
\doival{10.1017/xxxxx}

\title{\tabibert: A Large-Scale \modernbert\ Foundation Model and A Unified Benchmark for Turkish}

\begin{authgrp}
\author{Melikşah Türker$^{1,2}$, Asude Ebrar Kızıloğlu$^{1*}$, Onur Güngör$^{1}$, and Susan Üsküdarlı$^{1}$}
\renewcommand*{\thefootnote}{\fnsymbol{footnote}}
\footnotetext[1]{The work was conducted while the author was affiliated with the Department of Computer Engineering, Boğaziçi University, Istanbul, Türkiye. 
Current address: Technical University of Munich (TUM), Munich, Germany.}
\renewcommand*{\thefootnote}{\arabic{footnote}}

\affiliation{%
    \textsuperscript{1}Department of Computer Engineering, Boğaziçi University, Istanbul, Türkiye\\
    \textsuperscript{2}VNGRS, Istanbul, Türkiye\\
\email{meliksah.turker@std.bogazici.edu.tr, ebrar.kiziloglu@tum.de, 
onurgu@pt.bogazici.edu.tr, 
suzan.uskudarli@bogazici.edu.tr}
\corresp{Corresponding author: Melikşah Türker.}
}

\end{authgrp}

\history{(Received xx xxx xxx; revised xx xxx xxx; accepted xx xxx xxx)}

\begin{abstract}
The introduction of BERT established encoder-only transformer models as a foundational paradigm in natural language processing, motivating extensive research on architectural design and pretraining strategies.
\modernbert\ consolidated these improvements by integrating rotary positional embeddings, FlashAttention-based attention implementations, and refined normalization schemes to support longer contexts and more stable training.
Encoder-only models remain the standard tool for classification, tagging and retrieval, where contextual representations and low inference cost matter more than text generation, yet Turkish lacks a monolingual encoder trained from scratch that incorporates these advances.

This work introduces \tabibert, a monolingual Turkish encoder based on the \modernbert\ architecture that is trained from scratch on a large, carefully curated corpus.
\tabibert\ is pretrained on one trillion tokens sampled from an 86.58B-token multi-domain corpus comprising web text (72 per cent), scientific publications (19 per cent), source code (6 per cent), and mathematical content (0.3 per cent).
The model supports a context length of 8,192 tokens, sixteen times that of existing Turkish BERT models, and inherits the \modernbert\ architecture's efficiency at long context.
To support rigorous and reproducible evaluation, we introduce the \tabibench\ benchmark with 27 datasets across eight task categories with standardized splits and evaluation protocols.
Performance is summarized as a GLUE-style macro-average on a 0--100 scale.
\tabibert\ leads the Turkish models in five of eight task categories and \berturk, the previous best, in six of eight; the gains concentrate on question answering (+9.55 F1) and code retrieval (+2.41 NDCG@10), while the four short-text categories are near saturation.
Its average of 77.28 exceeds \berturk's 75.66; the multilingual \mmbert\ reaches 78.98 with twice the parameters and three times the training tokens, at 41 per cent more tokens per Turkish input.
We release model weights, training configurations, and evaluation code to provide a transparent and reproducible foundation for future Turkish encoder research.
\end{abstract}



\newpage
\maketitle

\section{Introduction}
\label{sec:introduction}

The architectural framework introduced by Bidirectional Encoder Representations from Transformers (BERT)~\citep{Devlin2019BERT} has been a foundational component of natural language understanding, enabling the induction of deeply contextualized linguistic features through self-supervised learning on large-scale corpora.
Its success produced a refined training recipe~\citep{Liu2019RoBERTa}, resource-efficient variants~\citep{Sanh2019DistilBERT,Jiang2020ConvBERT,Lan2020ALBERT,Clark2020ELECTRA}, domain-specialized models~\citep{Beltagy2019SciBERT,Huang2019ClinicalBERT,Lee2020BioBERT,Chalkidis2020LegalBERT}, multilingual encoders designed for cross-lingual transfer~\citep{Pires2019MultilingualBERT,wu2019betobentzbecassurprising,Conneau2020XLM-R}, and refinements to attention, positional encoding and normalization that relax its quadratic cost and 512-token limit~\citep{Beltagy2020Longformer,Dao2022FlashAttention,Su2024Roformer,Xiong2020LayerNorm,Zhang2019RootMeanSquare}.
These accumulated improvements were synthesized into ModernBERT, an updated encoder-only architecture that incorporates rotary positional embeddings, FlashAttention technology, alternating attention patterns, and refined pretraining procedures \citep{Warner2024ModernBERT}.
Its architecture supports context windows up to \num{8192} tokens, sixteen times BERT's original context length of 512, with improved computational efficiency that addresses limitations stemming from training instability and restricted sequence length \citep{Xiong2020LayerNorm, Beltagy2020Longformer}.

While these architectural advances have been widely adopted for high-resource languages, Turkish NLP remains primarily centered on earlier architectural paradigms.
BERTurk \citep{Schweter2020BERTurk} laid the foundation for transformer-based Turkish NLP as, to our knowledge, the first monolingual Turkish BERT model pretrained from scratch and publicly released. Its widespread adoption across diverse downstream tasks has made it the de facto standard for Turkish language understanding.
Subsequently, models such as ConvBERTurk and various ELECTRA adaptations have emerged on community platforms, although they frequently prioritized immediate availability over technical documentation.
Critical specifications regarding corpus composition, preprocessing pipelines, and systematic evaluation protocols are often underspecified, which impedes reproducibility and limits the ability of the research community to build systematically on prior work.

Recent large-scale multilingual efforts have incorporated Turkish alongside other languages.
Multilingual BERT (mBERT) \citep{Devlin2019BERT} and XLM-R \citep{Conneau2020XLM-R} share parameters and vocabulary across typologically diverse languages, which dilutes representational capacity for individual languages \citep{Conneau2020XLM-R, Chang2024Multilinguality}.
For morphologically rich languages, multilingual tokenizers produce inefficient segmentations, leading to increased token fertility, longer effective sequences, and reduced vocabulary utilization \citep{Rust2021HowGood}.

Although generative large language models now dominate the research agenda, encoder-only models remain the workhorse of a large class of applications.
Text classification, named entity recognition, part-of-speech tagging, semantic similarity and dense retrieval require contextual or pooled representations rather than autoregressive generation, and they are typically deployed in high-throughput or latency-sensitive settings where the cost per item matters \citep{Antoun2025ModernBERT}.
Encoders are also a component of generative systems rather than an alternative to them, serving as the retrieval stage of retrieval-augmented generation pipelines \citep{Lewis2020RAG}.
The economics favor them by more than an order of magnitude: \citet{Luccioni2024PowerHungry} measure task-specific text classification at a mean of 0.002 kWh per 1,000 queries against 0.047 kWh for text generation, and conclude that they `do not see convincing evidence for the necessity of [generative models'] deployment in contexts where tasks are well-defined'.
Nor is the quality gap what the difference in size suggests: with matched data and architecture, a fine-tuned 400M-parameter encoder outperforms a fine-tuned 1B-parameter decoder on MNLI (91.3 against 89.9), and its authors conclude that at 1B parameters or below `encoders will continue to outperform all others in their size range (and even ones above it)' \citep{Weller2026SeqVsSeq}.
A 149M-parameter Turkish encoder therefore answers a practical need: efficient, low-latency classification, tagging and retrieval for Turkish, which a multilingual encoder serves at a higher token cost and a generative model at a far higher inference cost.

\begin{table}[t]
\centering
\small
\caption{\label{tab:results-summary}Turkish BERT-based models across scale, training data, architectural features, and benchmark performance.}
\resizebox{0.65\linewidth}{!}{\begin{tabular}{
  l
  S[table-format=3.0]
  S[table-format=3.0]
  S[table-format=4.0]                 
  c
  c
  S[table-format=2.2]                 
}
\toprule
\multicolumn{1}{c}{\thead{Model}} &
\multicolumn{1}{c}{\thead{\# params \\(M)}} &
\multicolumn{1}{c}{\thead{Data size\\(GB)}} &
\multicolumn{1}{c}{\thead{Context\\length}} &
\multicolumn{1}{c}{\thead{Code and\\math}} &
\multicolumn{1}{c}{\thead{Flash\\Attention}} &
\multicolumn{1}{c}{\thead{\tabibench\\score}} \\
\midrule
   \midrule
   \tweetbert\          & 163    & 110   & 512       & \xmark        & \xmark    & 67.16      \\
   \ytubert\            & 111    & 75    & 512       & \xmark        & \xmark    & 71.95      \\
   \berturk\            & 110    & 35    & 512       & \xmark        & \xmark    & 75.66      \\
   \textbf{\tabibert}   & 149    & 426   & \B 8192   & \cmark        & \cmark    & \B 77.28 \\
\bottomrule
\end{tabular}}
\end{table}

This work identifies and addresses two distinct gaps in Turkish NLP research.
First, a \textit{modernization gap} exists: no monolingual Turkish encoder has been trained from scratch with the efficiency, stability and long-context improvements consolidated in \modernbert, so these properties have been unavailable to Turkish NLP.
Second, an \textit{evaluation gap} persists due to the lack of a standardized, reproducible benchmarking framework, hindering systematic comparison across model generations and architectural paradigms.

We address these gaps by introducing \tabibert, a monolingual Turkish encoder, based on the \modernbert\ architecture, and \tabibench, a unified evaluation benchmark with standardized protocols.
Across \tabibench, \tabibert\ leads the Turkish models in five of eight task categories and \berturk, the previous best, in six of eight; its overall average of 77.28 (a GLUE-style macro-average on a 0--100 scale) exceeds \berturk's 75.66.
\tabibert's technical specifications and results compared to prior Turkish BERT models are summarized in Table~\ref{tab:results-summary}.
The gains concentrate where the benchmark is not saturated: +9.55 F1 on question answering and +2.41 NDCG@10 on code retrieval, with smaller leads on academic understanding (+0.77 macro-F1) and information retrieval (+0.60 NDCG@10), whereas on the four short-text categories the leading Turkish models lie within 1 of one another.
Against the previous best result for each category individually, \tabibert\ is 1.48 higher on average, indicating consistent generalization across domains with a single general-purpose encoder.
Beyond accuracy, \tabibert\ supports a context length of 8,192 tokens, sixteen times that of prior Turkish BERT models, which cannot process such inputs at all; as a \modernbert-architecture model it also inherits that architecture's long-context efficiency, which \citet{Warner2024ModernBERT} report as up to 2.65$\times$ faster inference than the next-fastest encoder at 8,192 tokens.

For completeness, we also evaluate against \mmbert\ \citep{Marone2025mmBERT}, a recent large-scale multilingual encoder with \num{307}M parameters (against \tabibert's \num{149}M) and a \num{256000}-token vocabulary that achieves 78.98 on \tabibench.
Our work focuses on developing an efficient, monolingual encoder that incorporates modern architectural advances at a scale suitable for widespread research and practical deployment; Section~\ref{sec:mmbert-comparison} sets out when each of \tabibert, \berturk\ and \mmbert\ is the appropriate choice.

The main contributions of this work are as follows:

\begin{itemize}
    \item \tabibert~\citep{TabiBERT-HF}, a monolingual Turkish encoder pretrained from scratch on a corpus we composed and curated for this purpose: \num{86.58}B tokens (426 GB) of web text, academic articles, graduate theses, parliamentary transcripts, source code and mathematics, to our knowledge the first Turkish encoder corpus to include the last five.
    Pretrained for 1T tokens with the \modernbert\ architecture and recipe, it is the first monolingual Turkish encoder with an \num{8192}-token context and achieves state-of-the-art results among Turkish-specific encoders on five of eight benchmark categories.

    \item \tabibench~\citep{TabiBench}, a unified evaluation framework of 27 datasets across eight task categories with fixed public splits, a documented split-generation rule and an identical fine-tuning protocol for every compared model, covering domains such as code retrieval and academic text understanding that are absent from prior Turkish benchmarks.

    \item A reproducible release and two methodological findings: model weights, training configurations and evaluation code are public~\citep{TabiBERT}; and we document that \modernbert's published pretraining throughput depends on a pre-tokenization step not documented in its paper or codebase, and that token-classification comparisons across tokenizers require word-level rather than subword-level evaluation.
\end{itemize}

The remainder of this paper is organized as follows:
Section~\ref{sec:related-work} reviews prior work on encoder architectures, Turkish language models and evaluation benchmarks;
Section~\ref{sec:tabibert} describes the \tabibert\ model and the pretraining procedure;
Section~\ref{sec:tabibench} specifies the \tabibench\ evaluation framework;
Section~\ref{sec:evaluation} reports the evaluation results of \tabibert\ on the \tabibench\ benchmark;
Section~\ref{sec:discussion} discusses the results;
Section~\ref{sec:limitations} presents some limitations of our work; and
Section~\ref{sec:conclusions} concludes the paper and outlines directions for future work.
\section{Related Work}
\label{sec:related-work}

The introduction of Bidirectional Encoder Representations from Transformers (BERT)~\citep{Devlin2019BERT} was a significant advancement in encoder-only transformer models.
These models were trained on large English corpora and established strong performance baselines across diverse natural language processing tasks.
They provided a pretrained model that could be fine-tuned for various downstream tasks.

\subsection{Encoder-only Transformer Models and Architectural Evolution}
Following \bert, encoder-only transformer models were improved by various architectural innovations or training techniques, including changes to pretraining objectives, optimization strategies, and large-scale data and model scaling, leading to consistent performance gains across a wide range of downstream NLP tasks.
RoBERTa~\citep{Liu2019RoBERTa} scaled the corpus and training regime, ALBERT~\citep{Lan2020ALBERT} reduced model size through parameter sharing and factorized embeddings, ELECTRA~\citep{Clark2020ELECTRA} introduced the more sample-efficient replaced-token-detection objective, and the DeBERTa family~\citep{he2021debertav1,MicrosoftDeBERTaV2Repo,He2023DeBERTaV3} added disentangled attention and, in its third version, ELECTRA-style training.
Later work improved pretraining efficiency rather than the attention mechanism: MosaicBERT~\citep{Portes2023Mosaicbert} combined FlashAttention~\citep{Dao2022FlashAttention}, ALiBi~\citep{Press2021Train}, gated linear units~\citep{Shazeer2020Glu}, dynamic unpadding~\citep{Zeng2022Boosting} and low-precision normalization into a fast training recipe, RetroMAE~\citep{Xiao2022RetroMAE} redesigned masked autoencoding for dense retrieval, and FlexiBERT~\citep{Tuli2023FlexiBERT} searched heterogeneous encoder configurations with neural architecture search.
Renewed interest in encoder-only models has led to modernized successors to BERT. \modernbert~\citep{Warner2024ModernBERT} revisited the original BERT model, extending context length and updating architectural elements and design choices. NeoBERT~\citep{Breton2025NeoBERT} revisited the depth-to-width ratio and extended the context to \num{4096} tokens with deployment in mind, and OptiBERT~\citep{Dervishi2025TrainingOptiBERT} trained compute-optimal BERT-style models.

\subsection{Monolingual Turkish \bert\ Models}
The success of transformer-encoder models motivated a series of \bert\ models known as \berturk\ to be trained on Turkish data~\citep{Schweter2020BERTurk}.
Several variations of \bert, such as Distil\bert~\citep{Sanh2019DistilBERT}, \convberturk~\citep{Schweter2020BERTurkV2}, and ELECTRA~\citep{Clark2020ELECTRA}, were also trained on Turkish data by the same authors~\citep{Schweter2020BERTurk}.
These models were generally trained on a \num{35} GB corpus consisting of \num{4.4}B tokens drawn from the Turkish OSCAR corpus~\citep{OrtizSuarez2019OSCAR,Abadji2022OSCAR}, a Wikipedia dump,\footnote{An unspecified Wikipedia dump, presumed to date from approximately 2020.} various OPUS corpora~\citep{Tiedemann2012OPUS} and a corpus provided by Kemal Oflazer.
Some models, like \convberturk\ and ELECTRA, were also trained on the Turkish portion of the mC4 corpus~\citep{xue2021mt5massivelymultilingualpretrained,Schweter2020BERTurk}, which is a cleaned version of the public web crawl data of Common Crawl.
These models have been evaluated on various downstream tasks (such as part-of-speech tagging and named entity recognition) where they generally outperform their multilingual counterparts \citep{Schweter2020BERTurk,Conneau2020XLM-R}.
Despite their popularity, these models were limited in transparency as they were not products of academic publications.

Subsequently, two Turkish \bert\ models were introduced with academic publications.
\ytubert~\citep{Kesgin2023YTU} utilized an uncased Turkish tokenizer and was trained on a corpus of an unknown number of tokens reported to be around \num{75} GB in size, which consists of web crawls (OSCAR and mC4), Turkish Wikipedia, news and novels.
\tweetbert~\citep{Najafi2023TurkishBERTweet} was tailored for social media.
It also employed an uncased tokenizer, specifically designed for social media texts, by the addition of special tokens that can encode special information in tweets like hashtags, mentions, and URLs.
It was pretrained with a corpus of around \num{110} GB in uncompressed form, containing nearly 894 million tweets of at least 10 tokens each, that is, at least roughly \num{9}B tokens.

\subsection{Multilingual \bert\ Models}
The development of multilingual pretrained language models began with Multilingual BERT (mBERT)~\citep{Devlin2019BERT}, which extended the original BERT architecture to 104 languages by pretraining on a corpus obtained by concatenation of Wikipedia corpora.
Despite lacking explicit cross-lingual objectives, mBERT demonstrated surprising zero-shot cross-lingual transfer capabilities~\citep{Pires2019MultilingualBERT}.
XLM~\citep{Conneau2019Cross} worked on enhancing cross-lingual alignment and introduced Translation Language Modeling (TLM) that leverages parallel corpora.
XLM-RoBERTa~\citep{Conneau2020XLM-R} further scaled multilingual pretraining to \num{2.5} TB of Common Crawl data across 100 languages, achieving state-of-the-art performance using only masked language modeling without parallel data.
Later work varied the objective or the architecture: cross-lingual contrastive learning in InfoXLM~\citep{Chi2021InfoXLM}, back-translation masked language modeling in ERNIE-M~\citep{Ouyang2021ERNIE}, combined cross-lingual objectives in Unicoder~\citep{Huang2019Unicoder}, decoupled input and output embeddings in RemBERT~\citep{Chung2020Rethinking}, disentangled attention with ELECTRA-style pretraining in mDeBERTa~\citep{He2023DeBERTaV3}, and improved vocabulary allocation across languages in XLM-V~\citep{Liang2023XLMV}.

More recent work has further expanded the landscape of multilingual encoders.
EuroBERT~\citep{Boizard2025EuroBERT} introduced multilingual encoders covering European and several widely spoken global languages.
Similarly, \mmbert~\citep{Marone2025mmBERT}, building on \modernbert, incorporated long-context attention and efficient training strategies.
It demonstrated strong performance on classification, embedding, and retrieval tasks, particularly for low-resource languages, while largely preserving the inference efficiency of \modernbert\ despite its large vocabulary of \num{256000} tokens.

However, these broad multilingual approaches often struggle with the linguistic specificity required for morphologically complex languages, motivating the development of language-family-specific or language-specific models.

\subsubsection{Curse of Multilinguality}
A key challenge in multilingual modeling is the \emph{curse of multilinguality}, a phenomenon where adding more languages to a fixed-capacity model eventually degrades per-language performance~\citep{Conneau2020XLM-R}; it is attributed to negative interference between languages competing for shared parameters~\citep{Wang2020Negative}, has been observed across 10,000 models and 250 languages~\citep{Chang2024Multilinguality}, and is mitigated by modular adapters~\citep{Pfeiffer2022Lifting} or experts trained on language subsets~\citep{Blevins2024Breaking}.
These findings motivate the development of more targeted models that focus on either monolingual models or models that focus on related language families rather than attempting to cover all languages in a single model.

\subsection{\modernbert}
\modernbert\ is one of the recent and most significant bidirectional encoder-only transformer models. It upgrades the original BERT architecture by utilizing contemporary enhancements to support longer context lengths and improve efficiency and performance in natural language understanding tasks.
It was pretrained on a massive corpus of 2 trillion tokens of English text and code, enabling stronger generalization and suitability for both language and code understanding.
\modernbert\ introduces several architectural techniques aimed at improving efficiency, including Rotary Positional Embeddings (RoPE)~\citep{Su2024Roformer}, 
FlashAttention~\citep{Dao2022FlashAttention}, unpadding~\citep{Zeng2022Boosting}, also adopted by mGTE~\citep{Zhang2024Mgte}, GeGLU activations~\citep{Shazeer2020Glu}, and alternating attention~\citep{Team2024Gemma}.
Together, these design choices reduce memory usage and inference time compared to conventional encoder models, while preserving strong performance on downstream tasks.
These features make \modernbert\ effective for classification, retrieval, semantic search, and other NLU tasks, and it serves as an efficient alternative to older BERT variants in research and production settings.

\subsubsection{Language-Specific Variants}
The success of \modernbert\ has inspired several language-specific adaptations, demonstrating the effectiveness of specialized training over multilingual approaches.
For example, ModernGBERT is a family of German encoder models (134M, 1B parameters) trained from scratch, incorporating \modernbert's architectural innovations \citep{Wunderle2026ModernGBERT}.
Their work showed that dedicated encoders outperform encoders converted from German decoder-only LLMs via LLM2Vec~\citep{Behnamghader2024Llm2vec} in terms of both performance and parameter efficiency, validating the approach of language-specific model development.

\llmjpmodernbert\ is a similar \modernbert-based model trained on a large-scale Japanese corpus with \num{8192}-token context length \citep{Sugiura2025ModernBERTJapanese}.
It achieves good results on fill-mask evaluations but does not surpass existing baselines on downstream tasks, and its authors leave the cause open.

\subsubsection{Domain-Specific Variants}
\modernbert\ has been adapted to specialized domains that require long-context understanding: Clinical \modernbert\ was pretrained on biomedical literature, clinical notes and medical ontologies~\citep{Lee2025ClinicalModernBERT}, and BioClinical \modernbert\ continued pretraining on a multi-institution corpus of more than 53.5 billion tokens, outperforming existing biomedical and clinical encoders on four downstream tasks~\citep{Sounack2025BioClinicalModernBERT}.
In Japanese radiology report classification, \modernbert\ trained and ran faster than BERT at comparable in-distribution accuracy, although it generalized less well to an external set of reports~\citep{Yamagishi2025ModernBERTRadiology}; these adaptations motivate its use as a base for further domain- and language-adapted encoders.

\subsubsection{Architectural Analysis and Comparisons}
In a controlled study designed to isolate architectural effects from differences in pretraining data, \modernbert\ was compared with DeBERTaV3 when both models were pretrained on the same dataset.
The results indicate that while \modernbert\ offers faster training and inference speeds attributable to its architectural optimizations, the earlier DeBERTaV3 architecture maintained superior sample efficiency and overall benchmark performance under identical pretraining conditions~\citep{Antoun2025ModernBERT}.

\subsection{Turkish NLP Benchmarks}
Early evaluation of Turkish NLP models has predominantly relied on individual, task-specific datasets, developed independently for problems such as named entity recognition (NER), sentiment analysis, dependency parsing, or question answering.
Examples include Turkish NER datasets derived from Wikipedia~\citep{Rahimi2019WikiANN} and news~\citep{Tur2003MilliyetNER}, sentiment datasets collected from the web~\citep{Amasyali2018Sentiment}, and syntactic resources such as Universal Dependencies Turkish treebanks~\citep{UD25Turkish, UDTurkishIMST, UDTurkishBOUN}.
Even though these datasets have been useful in enabling steady and consistent progress on task-specific problems, they often differ substantially in annotation quality, domain coverage, and evaluation protocols, and many lack widely adopted, fixed train/validation/test splits, limiting the comparability of reported results across studies~\citep{Cakici2025CrossValidation}.
As transformer-based models became dominant, evaluation practices for Turkish largely continued to follow this fragmented paradigm.
Recent works frequently assembled their own evaluation datasets or introduced task-specific splits tailored to a single study.
For instance, \ytubert\ evaluated models on tasks such as sentiment analysis, news classification, mask prediction, and zero-shot classification using custom-constructed datasets and internally defined splits, rather than established benchmark splits~\citep{Kesgin2023YTU}.
\tweetbert\ focused on sentiment analysis and hate speech detection in social media, similarly relying on task-specific tweet datasets and cross-validation-based evaluation protocols rather than standardized benchmark test sets~\citep{Najafi2023TurkishBERTweet}.
Because each of these works evaluates its model on a self-selected set of datasets under its own splits or cross-validation folds, the reported results are not comparable across models: differences in data, splits and protocol confound any cross-study comparison, and a cross-validation study of Turkish sentiment analysis datasets and tools documents how strongly reported outcomes depend on exactly these choices~\citep{Cakici2025CrossValidation}.

In response to these limitations, several Turkish NLP benchmark efforts have emerged that aim to consolidate multiple tasks under unified evaluation frameworks.
Mukayese aggregated diverse Turkish NLP tasks, including language modeling, machine translation, NER, summarization, and text classification, into a single benchmark suite, providing common baselines~\citep{Safaya2022Mukayese}.
TrGLUE adapted the GLUE paradigm to Turkish natural language understanding tasks, though its scope remained largely limited to sentence-level NLU~\citep{Altinok2025TrGLUE}.
Cetvel, a large-scale LLM-oriented evaluation benchmark, covered a broader set of discriminative and generative tasks while emphasizing cultural and linguistic relevance for Turkish~\citep{kuisai2024cetvel}.
More recently, TR-MTEB  introduced a Turkish adaptation of the Massive Text Embedding Benchmark (MTEB), focusing on embedding and retrieval tasks~\citep{BaysanGungor2025TR-MTEB} by utilizing machine translation.

Despite recent advances, a systematic quality assessment of 17 widely used Turkish benchmark datasets against six criteria found that 70 per cent of them fail to meet the authors' heuristic standards for linguistic correctness, coherence and cultural appropriateness~\citep{Umutlu2025Evaluating}, underscoring the need for stricter dataset construction and validation standards.
Last but not least, most existing benchmarks focus on traditional NLP tasks, with limited coverage of modern NLP tasks such as academic text understanding, retrieval, and code-related tasks, which have become central to contemporary transformer and LLM evaluation.
No existing Turkish benchmark combines this task coverage with fixed, public splits and a uniform fine-tuning protocol across the compared encoders.

\section{TabiBERT: Model Architecture and Pretraining}
\label{sec:tabibert}

In this section, we present a comprehensive overview of the pretraining procedure of the \tabibert\ model, 
in terms of the pretraining dataset, tokenizer details, the pretraining objectives, and the model architecture.

\subsection{Pretraining Data}

We compiled a comprehensive dataset primarily consisting of Turkish monolingual text, supplemented with a small portion of English, coding and math data, to pretrain our model. 
To address significant size disparities and differences in quality across corpora and ensure balanced representation of different text types, we employed strategic oversampling of smaller but high-quality datasets.
The constituent corpora total \num{84.88}B raw tokens (sum of Table~\ref{tab:train_data}); weighting by sampling rate yields the \num{86.58}B-token pretraining corpus, of which \num{13} per cent is from non-Turkish sources.
Throughout the paper, including the abstract, all references to the size of the pretraining corpus include the effects of oversampling.
The details of each corpus are explained in the following subsections, and the training corpora statistics, as well as the sampling rates, are summarized in Table~\ref{tab:train_data}.
To our knowledge, this is the first Turkish encoder pretraining corpus to include academic articles, graduate theses, parliamentary transcripts, source code and mathematical text; prior Turkish encoders were trained on web crawls (OSCAR, mC4), Wikipedia, OPUS, news, novels or tweets.

\paragraph{Web corpora} We extracted the tur\_Latn subset of FineWeb-2 \citep{Penedo2025FineWeb2} as the main Turkish web corpus, with a knowledge cutoff date of April 2024.
We also included a sample of 5 million English web pages from the FineWeb-2 English dataset to provide English language context during pretraining. 
Moreover, we included a Turkish Wikipedia dataset comprising articles scraped from Wikipedia's Turkish-language pages in August 2024.
This dataset provides encyclopedic knowledge covering a wide range of topics written in a formal and informative style.

\paragraph{Scientific corpora} \label{sec:sci_cor} We used the Dergipark and Yöktez datasets from TURNA \citep{Uludogan2024TURNA}. 
Dergipark is a corpus of Turkish academic articles, and Yöktez is a corpus of Turkish graduate theses.
        
\paragraph{Book corpus} We used the Book Corpus dataset from TURNA \citep{Uludogan2024TURNA}, a compilation of Turkish fiction and non-fiction books. 

\paragraph{Bilkent Creative Writings} The \textit{Bilkent Creative Writings} \citep{Yilmaz2025BilkentWritings} corpus comprises creative writings produced by Bilkent University students in Turkish writing courses. 

\paragraph{ParlaMint-TR} The \textit{ParlaMint-TR} corpus originates from the CLARIN Flagship 
project,\footnote{\texttt{\href{https://www.clarin.eu/parlamint}{clarin.eu/parlamint}}} 
and contains transcripts of Turkish parliamentary debates within the broader European ParlaMint initiative. 

\paragraph{Code corpus} The \textit{Code} corpus consists of English code repositories, 
documentation and instructions obtained by randomly selecting a subset of \num{2.5}M examples from 
several coding corpora~\citep{CodeBagel2024, GitHubCode2022, PythonCodeAlpaca2023, TinyCodes2023}.

\paragraph{Math corpus} The \textit{Math} corpus comprises mathematical questions and solutions in English, 
collected from well-known academic and public datasets~\citep{Amini2019MathQA, HendrycksMath2021, Gao2024OmniMath, Yue2023Mammoth, Yu2023MetaMath, Hendrycks2024CompetitionMathNA}.

\begin{table}[htb!]
    \centering
    \small
    \caption{Training Corpora Statistics.}
    \setlength{\tabcolsep}{5pt} 
    \renewcommand{\arraystretch}{0.90} 
    \begin{tabular}{llS[table-format=8.0,round-mode=places,round-precision=0]S[table-format=5.2]S[table-format=1.0,round-mode=places,round-precision=0]}
    \toprule
    \textbf{Corpus} & \textbf{Type} & \textbf{\# docs} &  \textbf{\# tokens (B)} &  \textbf{Sampling rate} \\ \midrule
        FineWeb-2 Turkish  & Web & 88769907 &  56.041 & 1 \\
        FineWeb-2 English & Web & 5000000 &  5.676 & 1 \\ 
        Wikipedia-TR & Web & 604716 & 0.169 & 2 \\ 
        Dergipark  & Scientific & 334429 & 1.738 & 1 \\
        Yöktez  & Scientific & 475817 & 14.887 & 1 \\
        Books  & Literary & 5078 & 0.621 & 3 \\
        Bilkent Creative Writings  & Creative Text & 8457 &  0.007 & 3 \\
        ParlaMint-TR  & Dialogue & 1333 & 0.068 & 5 \\ 
        Code  & Code & 2500000 &  5.415 & 1 \\ 
        Math & Math & 817392 & 0.261 & 1 \\ \bottomrule
    \end{tabular}
    \label{tab:train_data}
\end{table}

\subsection{Tokenizer}
\label{sec:tokenizer}

In this work, we use the tokenizer developed for the \emph{Kumru} LLM~\citep{Turker2025Kumru}.

\paragraph{Specifications}
The Kumru tokenizer is a modern LLM tokenizer with a vocabulary of 50,176 tokens, trained with the byte-pair encoding (BPE)~\citep{Sennrich2016Neural} algorithm, and uses a modern pre-tokenization regular expression (RegEx) to split the text into contiguous character spans that correspond to natural token boundaries, such as words, numeric expressions, punctuation marks, and formatting symbols, before tokenization.
As a result of this pre-tokenization RegEx, symbols such as digits, punctuation marks, and special characters, including line breaks (\texttt{\textbackslash n}) and tab characters (\texttt{\textbackslash t}), are represented as single tokens.
This helps the model better understand and process code, mathematical equations, and page structure.

The tokenizer's design follows two recent studies.
Its vocabulary of \num{50176} lies at the upper end of the 33K--50K range that \citet{Ali2024Tokenizer} find compute-optimal for monolingual training, although that study covers English and four other European languages rather than Turkish; the pre-tokenization regular expression follows the recommendation of \citet{Dagan2024Getting}, who likewise use BPE.

\paragraph{Training dataset}
The tokenizer was trained on a dataset mixture that consists of 95 per cent Turkish and 5 per cent English, code, and math, a similar mixture to our model's pretraining data.
The dataset mixture was curated to support a wide range of language capabilities, including proficiency in Turkish and English, as well as the ability to handle code snippets and mathematical expressions.

\paragraph{Adaptation for TabiBERT}
To use a tokenizer developed for a decoder-only LLM with \tabibert, we added the special tokens [CLS] and [SEP] and modified the input processing template to make it compatible with BERT tasks.

\paragraph{Fertility}
\emph{Fertility}~\citep{Rust2021HowGood} measures the average number of tokens required to represent text and serves as an indicator of tokenizer representation efficiency.
Lower fertility is generally desirable for efficiency, but vocabulary size and fertility trade off against model size and compute~\citep{Ali2024Tokenizer}, and the same study cautions that fertility is `not always predictive of model downstream performance'.
We therefore treat fertility as an efficiency diagnostic rather than as a predictor of task accuracy.

We measure fertility values by tokenizing \num{1}M randomly selected web pages from our Turkish web corpus and report the relative fertility values compared to \tabibert\ in Figure~\ref{fig:tokenizer_fertility}.
We observe that:

\begin{itemize}
    \item \tabibert\ and \berturk, both cased, have comparable fertility despite \tabibert's much larger vocabulary; \tabibert's is about 3 per cent higher, which we attribute to its pre-tokenization rule reserving separate tokens for digits, punctuation, line breaks and tab characters, so that code, mathematical expressions and document structure are represented explicitly rather than merged into word pieces.

    \item \ytubert\ achieves the lowest fertility as a result of using an uncased tokenization strategy.

    \item As a result of being optimized for social media text, \tweetbert\ has a similar fertility value to \tabibert\ despite being uncased and having twice the vocabulary size.

    \item Despite its large vocabulary of \num{256000} tokens, the \mmbert\ tokenizer exhibits significantly higher fertility than Turkish-specific tokenizers.
    This reflects the limited representation of Turkish in large-scale multilingual training corpora such as FineWeb-2~\citep{Penedo2025FineWeb2}, where Turkish makes up only about 1.3 per cent of the retained data by bytes.\footnote{\texttt{\href{https://github.com/huggingface/fineweb-2/blob/main/fineweb2-language-distribution.csv}{github.com/huggingface/fineweb-2/blob/main/fineweb2-language-distribution.csv}}}
\end{itemize}

\begin{figure}[t]
\centering
\begin{tikzpicture}
    \begin{axis}[
        ybar,
        bar width=0.4cm,
        width=8cm,
        height=5cm,
        ylabel={Relative Fertility (per cent)},
        ylabel style={font=\small},
        symbolic x coords={YTU-Cosmos-BERT, BERTurk, TurkishBERTweet, mmBERT},
        xtick=data,
        axis x line=bottom,
        axis y line=left,
        clip=false,
        xticklabel style={
            font=\footnotesize,
            align=center,
            yshift=-5pt, 
        },
        xticklabels={
            YTU-Cosmos\\BERT\\(32K),
            BERTurk\\ \vphantom{ConvBERT}\\(32K),
            Turkish\\BERTweet\\(100K),
            mmBERT\\ \vphantom{ConvBERT}\\(256K)
        },
        ymin=-15,
        ymax=55,
        enlarge x limits=0.25,
        enlarge y limits={lower=0.2}, 
        grid=none,
        tick style={draw=none},
        axis line style={thick},
        extra y ticks={0},
        extra y tick style={
            grid=major,
            grid style={thick, dashed, gray!60}
        },
        extra y tick labels={},
    ]

    \addplot[fill=black, draw=black, bar shift=0pt] coordinates {
        (YTU-Cosmos-BERT, -9.4)
        (BERTurk, -2.6)
        (TurkishBERTweet, 1.2)
        (mmBERT, 41.0)
    };

    \node[font=\footnotesize\bfseries] at (axis cs:YTU-Cosmos-BERT,-9.4) [below, yshift=-1pt] {$-9.4$};
    \node[font=\footnotesize\bfseries] at (axis cs:BERTurk,-2.6) [below, yshift=-1pt] {$-2.6$};
    \node[font=\footnotesize\bfseries] at (axis cs:TurkishBERTweet,1.2) [above, yshift=1pt] {$+1.2$};
    \node[font=\footnotesize\bfseries] at (axis cs:mmBERT,41.0) [above, yshift=1pt] {$+41.0$};

    \end{axis}
\end{tikzpicture}
\caption{\label{fig:tokenizer_fertility}
Relative tokenizer efficiency, in per cent, on Turkish text across established Turkish BERT models and mmBERT, normalized to \tabibert\ (dashed baseline).
X-axis labels indicate models with their vocabulary sizes, while the Y-axis reports relative token counts required for text representation.
\tabibert's fertility is slightly higher than \berturk's despite its larger vocabulary, which we attribute to the pre-tokenization rule described in Section~\ref{sec:tokenizer}.
The multilingual mmBERT, despite its 256K-token vocabulary, exhibits 41.0 per cent higher fertility, reflecting greater subword fragmentation on Turkish text.
}
\end{figure}

\subsection{Pretraining Objectives}
For pretraining, we adopt the Masked Language Modeling (MLM) objective as in \modernbert, with a 30 per cent token masking rate.
We utilize the official \modernbert\ pretraining codebase and recipe to ensure methodological consistency and reproducibility.
The pretraining process is organized into three progressive phases, each exposing the model to 850B, 125B, and 25B tokens, respectively.
Hence, we scale the \modernbert\ pretraining recipe by 50 per cent with the aim of pretraining for \num{1}T tokens in total.

To best utilize the dataset and maximize training throughput, 
we pre-tokenize the entire pretraining corpus and chunk it into pieces that fit within the context length.
We perform optimization using the StableAdamW algorithm~\citep{Wortsman2023StableLowPrecision},
paired with a trapezoidal learning rate schedule~\citep{Xing2018Walk} and 1-sqrt decay~\citep{Hagele2024Scaling} for improved stability during large-scale training.
We pretrain our model for \num{117} hours on a cluster of 8$\times$ NVIDIA H100 GPUs, resulting in \num{1}T tokens being exposed to the model, that is, \num{11.55} epochs over the entire dataset.

\subsection{Model Architecture}
Our model is built on the \modernbert~\citep{Warner2024ModernBERT} architecture.
We use the \modernbert-base setup, with \num{22} hidden layers, 12 attention heads, an intermediate size of 1,152, a hidden size of 768, and GLU~\citep{Shazeer2020Glu} expansion to 2,304.
We set the vocabulary size to \num{50176} to match the tokenizer.

\tabibert\ shares every dimension that determines inference cost with \modernbert-base: 22 layers, hidden size 768, intermediate size 1,152, 12 attention heads, a 128-token local attention window with global attention every third layer, and \num{8192} positions.
The efficiency measurements of \citet{Warner2024ModernBERT} therefore transfer to it without re-measurement.
Inference throughput is fixed by architecture, input shape, kernel implementation, numerical precision and hardware rather than by weight values, and the one dimension in which the two models differ, the vocabulary, does not enter the computation: the input embedding is a table lookup, and the vocabulary-sized masked-language-modeling head is discarded for downstream tasks.
\citet{Warner2024ModernBERT} report that at 512 tokens \modernbert-base is comparable in speed to other base encoders and that at \num{8192} tokens it processes documents \num{2.65} times faster than the next-fastest encoder in their comparison.
For Turkish the more consequential fact is that the baselines of Section~\ref{sec:evaluation} cannot process \num{8192}-token inputs at all.
We have not run a dedicated inference benchmark of our own.

To facilitate replication and reproducibility, we release \tabibert\ model weights on Hugging Face~\citep{TabiBERT-HF},
in the hope that it will serve as a well-documented baseline incorporating \modernbert's architectural advances.

\section{TabiBench: A Standardized Turkish NLP Benchmark}
\label{sec:tabibench}

During the evaluation of \tabibert, we encountered significant challenges in identifying appropriate fine-tuning datasets for establishing reliable baseline comparisons.
A systematic review of commonly used Turkish NLP datasets revealed substantial fragmentation, which hampers the consistency and comparability of reported results~\citep{Cakici2025CrossValidation}: widely cited datasets suffer from inconsistent annotations, ambiguous labeling schemes, unclear guidelines or insufficient scale~\citep{Umutlu2025Evaluating}; standardized train/validation/test splits are lacking, so recent Turkish BERT studies rely on ad hoc evaluation datasets or custom splits~\citep{Kesgin2023YTU,Najafi2023TurkishBERTweet}; and existing benchmarks do not cover the task spectrum required to evaluate modern encoders.

Such challenges are common in lower-resource language NLP but are rarely addressed systematically.
Rather than compromise evaluation quality by using problematic datasets, 
we developed \tabibench\ as a standardized evaluation framework for Turkish NLP.
\tabibench\ is designed around three core principles: 
\begin{itemize}
    \item Quality over quantity: Systematic review of datasets for annotation consistency, label reliability, and task validity.
    \item Standardization: Fixed train-validation-test splits with explicit methodology for reproducibility and fair comparison.
    \item Task diversity: Inclusion of modern NLP tasks such as retrieval, code understanding, and domain specialization, as well as traditional NLU tasks.
\end{itemize}

By addressing the fragmentation and quality concerns evident in prior Turkish NLP benchmarks, 
\tabibench\ aims to enable more reliable, comprehensive, and future-facing evaluation of Turkish encoder models.
The remainder of this section describes the task categories and dataset coverage in detail, 
followed by \tabibench's release availability and the governance and provenance safeguards that keep its construction independent of \tabibert's development.

\subsection{Dataset Selection and Quality Assessment}
\label{sec:tabibench-selection}
We systematically review existing Turkish fine-tuning datasets, evaluating both dataset quality and task relevance.
To ensure our evaluation benchmark reflects realistic and meaningful downstream applications, datasets are selected based on three criteria:

\begin{itemize}
    \item Clear pattern-label correspondence: The relationship between input text and output labels must be unambiguous and learnable.
    \item Mutually exclusive labels: Each example must belong to exactly one class, ensuring well-defined classification boundaries.
    \item Documented annotation guidelines: Clear, consistently applied annotation protocols must be available to guarantee label reliability.
\end{itemize}

\subsection{Standardized Split Generation}
\label{sec:tabibench-splits}
Another issue encountered in Turkish NLP evaluation is the absence of standardized dataset splits.
Many existing datasets lack explicit train-validation-test divisions, leading researchers to create inconsistent splits that prevent fair cross-study comparisons.
We address this through a systematic split generation methodology applied uniformly across all datasets:

\begin{itemize}
    \item If a train-validation-test split is available, we use the existing divisions as provided.
    \item For datasets with a train-test split only, we create a validation split from the training set, sized to match the test set.
    \item In cases with a train-validation split, we reassign the validation set as the test split, then generate a new validation split from the training data following the approach above.
    \item In cases with a validation-test split, we split validation into train and validation sets in 80 and 20 per cent proportions, respectively.
    \item When only a single combined split is present, we partition the data into train, validation, and test sets in 70, 15, and 15 per cent proportions, respectively.
\end{itemize}

All splits use stratified sampling to preserve class distributions and ensure statistical validity.

\subsection{Task Categories and Dataset Coverage}
\label{sec:tabibench-tasks}

\tabibench\ encompasses 27 datasets spanning eight major task categories: 
text classification (4 datasets), token classification (4 datasets), semantic textual similarity (2 datasets), natural language inference (2 datasets), 
question answering (2 datasets), information retrieval (6 datasets), code retrieval (4 datasets), and academic understanding tasks (3 datasets).
Of these 27 datasets, 10 are natively Turkish, 2 are human-translated, 9 are machine-translated by prior work, and 6 are machine-translated by us using GPT-4.1 for the code retrieval and academic understanding categories.
Table~\ref{tab:tabibench-datasets} gives, for every dataset individually, its provenance (origin, translating agent where applicable, and source citation), task-level metric, label count, and train/validation/test split sizes,
so that the benchmark's composition can be read without cross-referencing the evaluation methodology (Section~\ref{sec:evaluation-methodology}) or the per-category results tables (Appendix~\ref{sec:appendix}); sources for each dataset are cited there rather than repeated below.
The information retrieval, code retrieval and academic understanding datasets require further description; for the other categories, Table~\ref{tab:tabibench-datasets} is self-contained.

\begin{itemize}
    \item Information retrieval covers six datasets of heterogeneous provenance, from general web search to scientific and financial documents: Scifact-TR, Fiqa-TR, NFCorpus-TR, and Quora-TR were translated by the TR-MTEB group using \texttt{aya-expanse-8b}; MsMarco-TR is a pre-existing Turkish machine translation of MS MARCO that TR-MTEB collected from prior work; and BiText is human-translated parallel text from the WMT16 En--Tr news-translation task together with SETIMES2 parallel text.

    \item Code retrieval datasets are translated from English using GPT-4.1 while preserving code integrity, so only the natural language components are translated: the problem descriptions of AppsRetrieval (Apps-TR), the docstrings of CodeSearchNet (CodeSearchNet-21K-TR, constructed by selecting \num{2500}, \num{500}, and \num{500} examples per programming language, over six languages, for training, validation, and testing), the questions of CosQA (CosQA-TR), and the questions and highest-voted answers of StackOverflow-QA (StackOverflowQA-TR).

    \item Academic understanding tasks assess model performance on specialized scholarly text.
    Two datasets are translated from English using GPT-4.1, while one dataset is natively constructed in Turkish:
    \begin{itemize}

        \item PubMedRCT-20K is a sentence-level classification dataset where sentences from scientific abstracts are labeled with rhetorical roles.
        A balanced subset of \num{10000} examples is selected using stratified sampling before translation; the resulting Turkish variant is PubMedRCT-10K-TR.

        \item SciCite is a three-class citation intent classification dataset, labeling citations as background, method, or result.
        Its Turkish variant, SciCite-TR, is created by translating the citation contexts while preserving the intent labels.

        \item Thesis-Abstract-Classification-11K is a natively constructed Turkish dataset derived from thesis abstracts collected from the National Thesis Center of Turkey.
        The task is to predict the field of study of a thesis abstract among 187 categories.
        A balanced subset with 60 samples per category is constructed, making this a challenging multi-class classification task due to the large label space, small number of samples per category, and specialized academic language.
    \end{itemize}
\end{itemize}

\begingroup
\footnotesize
\setlength{\tabcolsep}{2pt}
\renewcommand{\arraystretch}{1.15}
\begin{longtable}{@{}p{3.6cm}p{3cm}p{1.4cm}p{1.7cm}p{1.3cm}p{1.3cm}p{1.3cm}@{}}
    \caption{\tabibench\ per-dataset composition: provenance (origin and, where translated, translating agent, with source citation below each name) and evaluation setup (metric, label count, train/validation/test split sizes), for each of the 27 constituent datasets, grouped by the eight task categories of Section~\ref{sec:tabibench-tasks}. Counts: 10 native, 2 human-translated, 9 machine-translated by prior work, 6 machine-translated by us; 10+2+9+6=27.}
    \label{tab:tabibench-datasets} \\
    \toprule
    \textbf{Dataset} & \textbf{Provenance} & \textbf{Metric} & \textbf{\#Labels} & \textbf{Train} & \textbf{Val} & \textbf{Test} \\
    \midrule
    \endfirsthead
    \multicolumn{7}{l}{\textit{Table~\ref{tab:tabibench-datasets} continued}} \\
    \toprule
    \textbf{Dataset} & \textbf{Provenance} & \textbf{Metric} & \textbf{\#Labels} & \textbf{Train} & \textbf{Val} & \textbf{Test} \\
    \midrule
    \endhead
    \midrule
    \multicolumn{7}{r}{\textit{continued on next page}} \\
    \endfoot
    \bottomrule
    \endlastfoot
    \multicolumn{7}{l}{\textit{Text classification}} \\
    \midrule
    News Cat\newline{\scriptsize \citep{McemilgNewsCat}} & Native & Macro-F1 & 5 & \num{750} & \num{150} & \num{250} \\
    Product Reviews\newline{\scriptsize \citep{FthbrmnbyTurkishProductReviews}} & Native & Macro-F1 & 2 & \num{164615} & \num{35275} & \num{35275} \\
    Bil Tweet News\newline{\scriptsize \citep{Toraman2017EventPrediction,Toraman2021EventRetrieval}} & Native & Macro-F1 & 4 & \num{696} & \num{149} & \num{150} \\
    Gender Hate Speech Turkish\newline{\scriptsize \citep{Toraman2022HateSpeech,Sahinuc2023GenderBias}} & Native & Macro-F1 & 3 & \num{16002} & \num{1998} & \num{2000} \\
    \midrule
    \multicolumn{7}{l}{\textit{Token classification}} \\
    \midrule
    WikiNER\newline{\scriptsize \citep{Altinok2023Turkish}} & Native & Micro-F1 & 39 (BIO) & \num{17967} & \num{1000} & \num{1000} \\
    WikiANN-TR\newline{\scriptsize \citep{Rahimi2019WikiANN}} & Native & Micro-F1 & 7 (BIO) & \num{20000} & \num{10000} & \num{10000} \\
    PosUD-BOUN\newline{\scriptsize \citep{UDTurkishBOUN}} & Native & Micro-F1 & 17 (UD POS) & \num{7803} & \num{979} & \num{979} \\
    PosUD-IMST\newline{\scriptsize \citep{UDTurkishIMST}} & Native & Micro-F1 & 17 (UD POS) & \num{3435} & \num{1100} & \num{1100} \\
    \midrule
    \multicolumn{7}{l}{\textit{Semantic textual similarity}} \\
    \midrule
    SICK-TR\newline{\scriptsize \citep{Dehghan2025Turkish}} & MT, prior work & Pearson & N/A & \num{4500} & \num{500} & \num{4927} \\
    STSb-TR\newline{\scriptsize \citep{BekenFikri2021Summarization}} & MT, prior work & Pearson & N/A & \num{5749} & \num{1500} & \num{1379} \\
    \midrule
    \multicolumn{7}{l}{\textit{Natural language inference}} \\
    \midrule
    SNLI-TR\newline{\scriptsize \citep{BudurEtAl2020Data}} & MT, prior work & Macro-F1 & 3 & \num{549367} & \num{9842} & \num{9824} \\
    MultiNLI-TR\newline{\scriptsize \citep{BudurEtAl2020Data}} & MT, prior work & Macro-F1 & 3 & \num{392702} & \num{4892} & \num{4923} \\
    \midrule
    \multicolumn{7}{l}{\textit{Question answering}} \\
    \midrule
    TQuAD\newline{\scriptsize \citep{ErdometoTQuad2}} & Native & F1 (span) & N/A & \num{11803} & \num{2418} & \num{2520} \\
    XQuAD\newline{\scriptsize \citep{Artetxe2019CrossLingual}} & Human-translated & F1 (span) & N/A & \num{833} & \num{178} & \num{179} \\
    \midrule
    \multicolumn{7}{l}{\textit{Information retrieval}} \\
    \midrule
    BiText\newline{\scriptsize \citep{Bojar2016WMT16}} & Human-translated & NDCG@10 & N/A & \num{205756} & \num{1001} & \num{3000} \\
    MsMarco-TR\newline{\scriptsize \citep{BaysanGungor2025TR-MTEB,Kesgin2023YTU,Muennighoff2023MTEB}} & MT, prior work (pre-existing translation collected by TR-MTEB, attributed to \citet{Kesgin2023YTU}) & NDCG@10 & N/A & \num{253332} & \num{31540} & \num{31692} \\
    Scifact-TR\newline{\scriptsize \citep{BaysanGungor2025TR-MTEB,Muennighoff2023MTEB}} & MT, prior work (\texttt{aya-expanse-8b}) & NDCG@10 & N/A & \num{580} & \num{339} & \num{339} \\
    Fiqa-TR\newline{\scriptsize \citep{BaysanGungor2025TR-MTEB,Muennighoff2023MTEB}} & MT, prior work (\texttt{aya-expanse-8b}) & NDCG@10 & N/A & \num{14166} & \num{1238} & \num{1706} \\
    NFCorpus-TR\newline{\scriptsize \citep{BaysanGungor2025TR-MTEB,Muennighoff2023MTEB}} & MT, prior work (\texttt{aya-expanse-8b}) & NDCG@10 & N/A & \num{110575} & \num{11385} & \num{12334} \\
    Quora-TR\newline{\scriptsize \citep{BaysanGungor2025TR-MTEB,Muennighoff2023MTEB}} & MT, prior work (\texttt{aya-expanse-8b}) & NDCG@10 & N/A & \num{6100} & \num{1526} & \num{15675} \\
    \midrule
    \multicolumn{7}{l}{\textit{Code retrieval}} \\
    \midrule
    Apps-TR\newline{\scriptsize \citep{Hendrycks2021Measuring}} & MT, ours: GPT-4.1 & NDCG@10 & N/A & \num{1235} & \num{3765} & \num{3770} \\
    CosQA-TR\newline{\scriptsize \citep{Huang2021CosQA}} & MT, ours: GPT-4.1 & NDCG@10 & N/A & \num{19604} & \num{500} & \num{500} \\
    StackOverflowQA-TR\newline{\scriptsize \citep{Li2025Coir}} & MT, ours: GPT-4.1 & NDCG@10 & N/A & \num{13951} & \num{3986} & \num{1994} \\
    CodeSearchNet-21K-TR\newline{\scriptsize \citep{Husain2019CodeSearchNet}} & MT, ours: GPT-4.1 & NDCG@10 & N/A & \num{15000} & \num{3000} & \num{3000} \\
    \midrule
    \multicolumn{7}{l}{\textit{Academic understanding}} \\
    \midrule
    PubMedRCT-10K-TR\newline{\scriptsize \citep{Dernoncourt2017Pubmed}} & MT, ours: GPT-4.1 & Macro-F1 & 5 & \num{7000} & \num{1500} & \num{1500} \\
    SciCite-TR\newline{\scriptsize \citep{Cohan2019Structural}} & MT, ours: GPT-4.1 & Macro-F1 & 3 & \num{8194} & \num{916} & \num{1859} \\
    Thesis-Abstract-Classification-11K\newline{\scriptsize \citep{TurkishThesis2025Ertugrul}} & Native & Macro-F1 & 187 & \num{7854} & \num{1683} & \num{1683} \\
\end{longtable}
\endgroup

\subsection{Impact and Availability}

\tabibench\ addresses an evaluation-infrastructure gap common to lower-resource languages that lack standardized benchmarks comparable to GLUE~\citep{Wang2018GLUE} and SuperGLUE~\citep{Wang2019SuperGLUE}, and its construction can serve as a template for them.
We release it on Hugging Face~\citep{TabiBench}, subject to the per-dataset redistribution status documented in Section~\ref{sec:tabibench-governance}, to establish evaluation practice for Turkish NLP, enable direct cross-model comparison and provide reproducible infrastructure for measuring progress.

\subsection{Benchmark Governance and Provenance}
\label{sec:tabibench-governance}

Because \tabibench\ is compiled by the authors of this paper and is subsequently used to evaluate \tabibert\ (Section~\ref{sec:evaluation}),
we document its provenance, licensing status, and the safeguards that keep the benchmark's construction independent of \tabibert's development,
so that \tabibench\ can be audited and reused on the same terms as any external benchmark.
As detailed below, we report these safeguards as-performed, not as a preregistration: \tabibench's construction and \tabibert's evaluation
were concurrent efforts by the same authors (Section~\ref{sec:tabibench}), not sequential or blinded ones, and no snapshot of the dataset
splits was publicly time-stamped or frozen before evaluation began. The independence we describe is a property of the selection and
evaluation procedures themselves, not of the order or isolation in which the two efforts occurred.

\paragraph{Scope of authorship}
\tabibench\ is an aggregation and standardization effort, not a from-scratch data collection effort.
Of its 27 constituent datasets, 21 are pre-existing resources that we sourced, quality-screened, 
and re-split under a common protocol (Sections~\ref{sec:tabibench-selection}--\ref{sec:tabibench-splits}).
This group of 21 includes the six retrieval datasets assembled by TR-MTEB~\citep{BaysanGungor2025TR-MTEB}, 
and one dataset, Thesis-Abstract-Classification-11K, that we curated as a class-balanced subset of an existing, 
publicly released Turkish thesis-abstract corpus~\citep{TurkishThesis2025Ertugrul} rather than translated.
The remaining 6 datasets, all four code retrieval datasets and two of the three academic understanding datasets, 
are English-language datasets that we translated into Turkish ourselves using GPT-4.1, 
translating only natural language components and leaving code and labels unmodified (Section~\ref{sec:tabibench-tasks}).
In every case, our contribution is dataset selection, quality vetting, translation where applicable, 
and standardized splitting; we are not the original creators of the underlying text or annotations, and we make no claim of authorship over source content.

\paragraph{Selection and quality review}
Constituent datasets were identified through a systematic review of existing Turkish NLP resources 
and retained only if they met the three criteria defined in Section~\ref{sec:tabibench-selection}; 
datasets that failed these criteria, or that were judged too small, ambiguously labeled, 
or inconsistently annotated during this review, were excluded rather than repaired.
Splits were then regenerated for every dataset under the deterministic procedure described in Section~\ref{sec:tabibench-splits}.
No dataset's inclusion, exclusion, or split assignment was informed by \tabibert's performance: 
the selection criteria are text- and annotation-level properties of each candidate dataset, not of any model's score on it.

\paragraph{Licensing and redistribution status}
\tabibench\ does not carry a single, blanket license.
Because it aggregates resources from many independent providers under heterogeneous terms, undeclared for a substantial share of sources, we deliberately do not attach one project-wide license to the aggregate release; instead, each constituent
dataset retains and is redistributed under its own original license or terms of use, tracked individually.
Table~\ref{tab:benchmark_governance} (Appendix~\ref{sec:appendix}) reports, for each constituent dataset,
its original provider, the license or terms we were able to locate, and its resulting redistribution status within \tabibench.
In summary: several sources carry an explicit open license (e.g., CC~BY-SA~4.0, MIT), and a smaller number carry a non-commercial or
share-alike license that restricts downstream reuse (e.g., CC~BY-NC-SA~4.0).
\tabibench\ is released on Hugging Face~\citep{TabiBench} on the per-dataset basis recorded in Table~\ref{tab:benchmark_governance}:
openly licensed and unrestricted sources are redistributed directly, and sources with no stated license are provided in the same form
and under the same access conditions as their original release; the dataset cards carry no license tags of our own, and the status recorded in the table is the authoritative statement of each dataset's terms.
One dataset was excluded on these grounds during the revision.
We had translated MedNLI~\citep{Romanov2018Lessons}, whose source corpus is distributed through PhysioNet under a credentialed-access agreement,
from a copy obtained from a public repository that carries no license, and we could not establish that our use of that copy, or a public release of its Turkish translation, was authorized.
MedNLI-TR was therefore removed from the benchmark, from all reported results and from the public release, for reasons of data access and not of results; we intend to apply for credentialed access and, if it is granted, to include a translation in a future version of \tabibench.
For the sources where we were unable to locate a stated license, we report that status explicitly rather than presuming permissiveness; these differ from MedNLI in that their original distribution is open and not access-controlled.
All 27 datasets were obtained and used for research purposes only, and our derived versions, the standardized splits and the six translations, are redistributed under the terms of their respective sources, where stated, rather than under a new license of our own.
We did not conduct an independent legal review of these terms, and we encourage users with license-sensitive use cases to verify terms against the original sources cited in Table~\ref{tab:benchmark_governance} before redistributing or building on \tabibench.

\paragraph{Independence from model validation}
Because the same authors constructed \tabibench\ and developed \tabibert, we take further steps to keep the benchmark's evaluative role independent of any single model's design.
Dataset selection, quality filtering, and split generation, described above, were fixed based on properties of the candidate datasets themselves, independent of \tabibert's architecture or training configuration.
Every model compared in this paper, including \tabibert, is hyperparameter-tuned and fine-tuned under the identical fair comparison protocol of Section~\ref{sec:evaluation-methodology}, which differs between models only in the maximum context length, capped at each model's supported limit: all five models are searched over the same hyperparameter grid (Table~\ref{tab:hp_space}); the best configuration for each model-task pair is selected by validation-set performance only, with the held-out test split used exactly once per model, for final reporting, and never for model or checkpoint selection; fine-tuning uses a single fixed random seed (25) across all models; and the two uncased baselines (\ytubert, \tweetbert) receive the same Turkish-aware lowercasing accommodation, applied uniformly rather than tuned per model.
Per-category scores are computed by a single, pre-specified aggregation rule applied uniformly across all models and all eight task categories, a test-size-weighted average across the constituent datasets of a category (Section~\ref{sec:evaluation-methodology}); this rule was fixed as part of the evaluation protocol before any test score was inspected and was not revisited after results were observed.
\tabibench\ is also released independently of this paper, with its full construction methodology, splits, and processing scripts made public, so that its quality claims and \tabibert's results on it can be independently audited or contested, and so that future work can reuse it to evaluate other models.

\paragraph{Full disclosure of results, including where \tabibert\ trails}
We report \tabibert's results in full, including the categories where it trails the Turkish baselines (text classification, semantic textual similarity and token classification) and the overall average on which \mmbert\ exceeds it (Table~\ref{tab:sft-overall}, Sections~\ref{sec:results} and~\ref{sec:mmbert-comparison}), as evidence that \tabibench's construction and aggregation were not adjusted to secure a uniform advantage for \tabibert.

\paragraph{Known residual risks}
The machine-translated datasets inherit translation-quality risk; for the six we translated ourselves it is quantified by the adequacy audit in Section~\ref{sec:translation-limitations}.

\section{Evaluation}
\label{sec:evaluation}

This section presents a comprehensive evaluation of \tabibert\ across diverse Turkish NLP tasks.
We first introduce the baseline models used for comparison, 
then describe our evaluation methodology, and present results demonstrating \tabibert's strengths and limitations.

\subsection{Models Used for Comparison}

We compare \tabibert\ against three monolingual Turkish encoder-only models based on the BERT-Base configuration: \berturk~\citep{Schweter2020BERTurk}, \ytubert~\citep{Kesgin2023YTU}, and \tweetbert~\citep{Najafi2023TurkishBERTweet}.
\berturk, though not peer-reviewed, is long-standing and widely adopted in Turkish NLP; \tweetbert\ is peer-reviewed and \ytubert\ is described in a detailed preprint; both are well documented.
All three are pretrained on large, diverse Turkish corpora, constituting strong representative baselines.

Additionally, we include the multilingual \mmbert~\citep{Marone2025mmBERT}, a ModernBERT-based encoder with 307M parameters covering approximately \num{1800} languages.
\mmbert\ differs substantially from \tabibert\ in scale and training scope (3$\times$ training tokens) and is not directly comparable.
We include it as an informative reference point contextualizing Turkish-focused models within large-scale multilingual settings.

\subsection{Evaluation Methodology}
\label{sec:evaluation-methodology}

\paragraph{Hyperparameter tuning}
We perform a systematic hyperparameter search for all model-task pairs to ensure fair comparisons.
The search space, informed by prior work~\citep{He2023DeBERTaV3, Warner2024ModernBERT, Breton2025NeoBERT}, is presented in Table~\ref{tab:hp_space}.
All models train for up to 10 epochs with early stopping (patience=2) based on validation loss.
Best configurations are selected via validation performance for the final test evaluation.

\begin{table}[t!]
    \centering
    \caption{Hyperparameter Search Space.}
    \label{tab:hp_space}
    \begin{tabular}{ll}
    \toprule
    \textbf{Parameter} & \textbf{Values} \\
    \midrule
    Learning Rate & \num{5e-6}, \num{1e-5}, \num{2e-5}, \num{3e-5} \\
    Weight Decay & \num{1e-5}, \num{1e-6} \\
    Batch Size & 16, 32 \\
    Epoch & 10, with early stopping \\
    \bottomrule
    \end{tabular}
\end{table}

\paragraph{Fine-tuning procedure}
We use optimal fine-tuning parameters identified during hyperparameter tuning, with a linear warmup-decay scheduler (6 per cent warmup, decay to 0.02$\times$ learning rate), mixed precision training, and random seed 25 for reproducibility.
Configurations follow ModernBERT's fine-tuning examples.\footnote{\texttt{\href{https://github.com/AnswerDotAI/ModernBERT/blob/main/yamls/finetuning/hf-bert-base-uncased.yaml}{github.com/AnswerDotAI/ModernBERT/yamls/finetuning/}}}

\paragraph{Fair comparison protocol} Note that we perform hyperparameter tuning followed by fine-tuning for \emph{all} models in our evaluation using \tabibench, not just \tabibert.
This ensures all models are evaluated at their best performance on these tasks, making comparisons attributable to architectural differences and explicit design decisions rather than hyperparameter selection discrepancies.

\paragraph{Context length}
We compute the 95th percentile sequence length for each dataset and set the maximum context length to $\min(\text{model context limit}, \text{dataset 95th percentile})$.
If the 95th percentile exceeds the model's maximum context length, we use the model limit.
This design choice enables efficient use of available computational resources while maximizing context utilization.
\berturk, \ytubert, and \tweetbert\ are limited to 512 tokens, whereas \tabibert\ and \mmbert\ support input lengths of up to \num{8192} tokens, enabling the use of longer contexts when available.
Appendix~\ref{sec:context-ablation} quantifies how much of \tabibert's performance gain stems from this longer context versus modeling improvements, 
with an ablation of \tabibert's performance on question answering tasks, where the largest gain is observed.

\paragraph{Training objectives}
Task-specific objectives are used across the benchmark:
cross-entropy is utilized for text classification, NLI, and token classification;
Mean Squared Error (MSE) loss is applied for semantic similarity regression;
span prediction via cross-entropy is implemented for question answering; and
contrastive learning is used for retrieval tasks.

\paragraph{Uncased models} For uncased models (\ytubert, \tweetbert), we apply Turkish-aware lowercasing (\texttt{text.replace("I", "ı").lower()}).

\paragraph{Evaluation metrics}
Task-appropriate metrics are adopted for performance assessment:
macro-averaged F1 is used for text classification and NLI;
micro-averaged F1 at the word level is used for token classification;
the Pearson correlation coefficient is used for semantic textual similarity (STS);
the F1 score is used for question answering; and
Normalized Discounted Cumulative Gain at rank 10 (NDCG@10) is implemented for retrieval tasks.

\paragraph{Aggregation of results} All metrics are reported on a 0--100 scale. For each task category, we report a single score via a weighted average across datasets, with weights proportional to test set size (ranging from 150 to 35K examples).
This ensures that the aggregate score reflects performance across all test instances rather than giving equal weight to small and large datasets, which would overrepresent performance on smaller datasets.
The detailed results per dataset for each task category are presented in Appendix~\ref{sec:appendix}.

\subsection{Results}
\label{sec:results}
\begin{figure*}[t]
\centering
\pgfplotsset{
    grayscale style/.style={
        legend image code/.code={
            \draw[##1] (0cm,-0.1cm) rectangle (0.3cm,0.1cm);
        }
    }
}
\begin{tikzpicture}
\begin{axis}[
    ybar,
    width=0.95\textwidth,
    height=6.5cm,
    bar width=4pt,
    ylabel={Score (0--100)},
    ylabel style={font=\small},
    ymin=0, ymax=115,
    ytick={0,20,40,60,80,100},
    xtick=data,
    xticklabels={Text Clf, Tok. Clf, STS, NLI, QA, Acad. Und., Info Retr., Code Retr.},
    xticklabel style={font=\small},
    ymajorgrids=true,
    grid style=dashed,
    enlarge x limits=0.08,
    legend style={
        at={(0.5,-0.22)},
        anchor=north,
        legend columns=5,
        font=\footnotesize,
        draw=none
    },
    legend image code/.code={
        \draw[draw=black, fill=#1, line width=0.6pt]
            (0cm,-0.1cm) rectangle (0.3cm,0.1cm);
    }
]

\addplot[fill=black!15, draw=black, line width=0.6pt] coordinates {
    (0,79.71) (1,92.02) (2,75.86) (3,79.10)
    (4,38.13) (5,60.60) (6,68.40) (7,43.49)
};

\addplot[fill=black!30, draw=black, line width=0.6pt] coordinates {
    (0,84.25) (1,93.60) (2,84.68) (3,84.16)
    (4,31.50) (5,69.29) (6,74.29) (7,53.80)
};

\addplot[fill=black!45, draw=black, line width=0.6pt] coordinates {
    (0,83.42) (1,93.67) (2,85.33) (3,84.33)
    (4,60.16) (5,69.00) (6,74.84) (7,54.54)
};

\addplot[fill=black!80, draw=black, line width=0.8pt] coordinates {
    (0,83.44) (1,93.42) (2,84.74) (3,84.51)
    (4,69.71) (5,70.06) (6,75.44) (7,56.95)
};

\addplot[pattern=crosshatch, pattern color=black, draw=black, line width=0.6pt] coordinates {
    (0,82.54) (1,93.81) (2,87.05) (3,84.38)
    (4,71.47) (5,70.34) (6,76.20) (7,66.02)
};

\legend{TurkishBERTweet, YTU-Cosmos-BERT, BERTurk, TabiBERT, mmBERT}
\end{axis}
\end{tikzpicture}
\caption{Task-specific performance of models on \tabibench\ by category: Text\&NLI\&Academic (macro-F1), Token Clf (micro-F1), STS (Pearson), QA (F1), and Information Retrieval\&Code Retrieval (NDCG@10). \tabibert\ consistently performs strongly across all tasks.}
\label{fig:performance_comparison}
\end{figure*}

Table~\ref{tab:sft-overall} presents the complete results; Figure~\ref{fig:performance_comparison} visualizes the per-category scores.

\begin{table}[t!]
   \centering
   \caption{Comparison of downstream task performance across all evaluated models. 
   For each column, the highest score among the Turkish models (excluding the multilingual \mmbert) is shown in bold.
   The evaluation metric used for each task type is also displayed in the column headers.
   \tabibert\ achieves the highest score in five of the eight task categories among Turkish models; \berturk\ leads two (token classification, STS) and \ytubert\ one (text classification).
   The last two rows give, for each category, the best score among the three Turkish baselines and \tabibert's margin over it; the previous best is \ytubert\ for text classification and academic understanding and \berturk\ for the other six categories. In these two rows the Total Avg column holds the mean of the eight per-category previous bests and \tabibert's margin over that mean.}
   \label{tab:sft-overall}
   \sisetup{
      detect-weight = true,
      detect-inline-weight = math,
      retain-explicit-plus = true
  }
  \resizebox{\linewidth}{!}{
   \begin{tabular}{
       l
       S[table-format=3.0]
       S[table-format=2.2]
       S[table-format=2.2]
       S[table-format=2.2]
       S[table-format=2.2]
       S[table-format=2.2]
       S[table-format=2.2]
       S[table-format=2.2]
       S[table-format=2.2]
       S[table-format=2.2]
   }
   \toprule
    \textbf{Model}  
        & \textbf{\# of params}
        & \textbf{Text clf.} 
        & \textbf{Token clf.} 
        & \textbf{STS} 
        & \textbf{NLI} 
        & \textbf{QA} 
        & \textbf{Academic understanding} 
        & \textbf{Information retrieval} 
        & \textbf{Code retrieval} 
        & \textbf{Total avg.}  \\
        & \multicolumn{1}{c}{M} 
        & \multicolumn{1}{c}{macro-F1} 
        & \multicolumn{1}{c}{micro-F1} 
        & \multicolumn{1}{c}{Pearson} 
        & \multicolumn{1}{c}{macro-F1} 
        & \multicolumn{1}{c}{F1} 
        & \multicolumn{1}{c}{macro-F1} 
        & \multicolumn{1}{c}{NDCG@10} 
        & \multicolumn{1}{c}{NDCG@10} 
        & \multicolumn{1}{c}{(\tabibench)} \\
   \midrule
   \tweetbert\  &163        & 79.71               & 92.02          & 75.86           & 79.10           & 38.13          & 60.60           & 68.40              & 43.49                   & 67.16      \\ 
   \ytubert\    &111        & {\B 84.25}          & 93.60          & 84.68           & 84.16           & 31.50          & 69.29           & 74.29              & 53.80                   & 71.95      \\
   \berturk\    &110        & 83.42               & \B 93.67       &  \B 85.33       & 84.33           & 60.16          & 69.00           & 74.84              & 54.54                   & 75.66      \\
   \textbf{\tabibert} &149  & 83.44               & 93.42          & 84.74           & \B 84.51        & \B 69.71       & \B 70.06        & \B 75.44           & \B 56.95                & \B 77.28 \\ \midrule
   \mmbert\     &307        & 82.54               & 93.81          & 87.05           & 84.38           & 71.47          & 70.34           & 76.20              & 66.02                   & 78.98      \\ \midrule
   Previous best (Turkish) & {} & 84.25 & 93.67 & 85.33 & 84.33 & 60.16 & 69.29 & 74.84 & 54.54 & 75.80 \\
   $\Delta$ \tabibert\ & {} & -0.81 & -0.25 & -0.59 & +0.18 & +9.55 & +0.77 & +0.60 & +2.41 & +1.48 \\ \bottomrule
   \end{tabular}
   }

\end{table}

\subsubsection{Overall Performance}

\tabibert\ establishes the highest average among monolingual Turkish models (77.28), followed by \berturk\ (75.66) and \ytubert\ (71.95).
The multilingual \mmbert\ achieves 78.98, reflecting its substantially greater resources: 2.1$\times$ parameters (307M against 149M) and 3$\times$ training tokens (3T vs. 1T).

Against the best Turkish baseline in each category (the last two rows of Table~\ref{tab:sft-overall}), \tabibert\ leads in five of the eight; in those five it exceeds the closest competitor by 2.70 on average, with the large gains in question answering (+9.55 F1) and code retrieval (+2.41 NDCG@10) over \berturk, while the leads on NLI (+0.18 macro-F1), information retrieval (+0.60 NDCG@10) and academic understanding (+0.77 macro-F1) are below 1.
Where it trails, on text classification ($-0.81$ macro-F1), STS ($-0.59$ Pearson) and token classification ($-0.25$ micro-F1), the deficits are likewise below 1.
Its mean of 77.28 exceeds the mean of the per-category previous bests, 75.80, by 1.48.

The pattern is therefore a decomposition rather than a uniform advantage.
On the four short-text categories (text classification, token classification, STS, NLI) the leading Turkish models lie within 1 of one another and the aggregate is near saturation for base-sized encoders, whereas question answering and code retrieval separate the models clearly on their respective metrics (9.55 F1 and 2.41 NDCG@10).
The context-length ablation (Appendix~\ref{sec:context-ablation}) shows that most of the question answering gain persists when \tabibert\ is capped at 512 tokens, so it is a model effect rather than a context-window effect alone.

\subsubsection{Performance by Task Category}

\paragraph{Text understanding tasks}

On text classification, results converge among top models (except \tweetbert). 
\ytubert\ leads (84.25), followed closely by \tabibert\ (83.44) and \berturk\ (83.42). Margins under 1 macro-F1 suggest approaching saturation.

On token classification, all Turkish models exceed 92 micro-F1. \berturk\ (93.67), \ytubert\ (93.60), and \tabibert\ (93.42) cluster within 0.25 micro-F1, indicating strong performance across models with limited improvement headroom.

On semantic textual similarity, top models (excluding \tweetbert) score a Pearson correlation of 84.68--85.33. 
\berturk\ leads (85.33), \tabibert\ follows (84.74), then \ytubert\ (84.68).

On natural language inference, \tabibert\ achieves the highest Turkish model score (84.51), marginally exceeding \berturk\ (84.33) and \ytubert\ (84.16).

\paragraph{Question answering}
\tabibert\ shows the largest improvement over Turkish baselines: 69.71 F1 versus \berturk's 60.16 (+9.55 F1, about 16 per cent relative gain).
Of the two datasets in this benchmark, only TQuAD's 95th-percentile length (\num{1024} tokens) exceeds the \num{512}-token limit of \berturk, \ytubert, and \tweetbert, forcing these three baselines to truncate its longer passages, 
whereas XQuAD's 95th percentile (\num{384} tokens) stays under this limit and is unaffected; 
\tabibert\ and \mmbert\ instead share an \num{8192}-token window, processing TQuAD without truncation.
A context-length ablation (Appendix~\ref{sec:context-ablation}) decomposes \tabibert's TQuAD gain into these two components: capping \tabibert\ at 512 tokens on TQuAD still yields 70.31 F1, well above \berturk's 63.30, 
indicating that context length accounts for only about 2 of the 9.04 F1, while the remaining roughly 7 F1 reflect a genuine modeling advantage independent of context.

\paragraph{Academic understanding tasks}
\tabibert\ leads Turkish models (70.06), exceeding \ytubert\ (69.29) and \berturk\ (69.00).
It is particularly strong on Thesis-Abstract-Classification-11K (187-class task with specialized language and limited per-class data), 
where pretraining on scientific articles and theses provides a domain advantage, in addition to the impact of its long-context support.

\paragraph{Information retrieval}
\tabibert\ achieves 75.44 NDCG@10 across the six information retrieval datasets, surpassing \berturk\ (74.84).
The gain is modest and we do not attribute it to any single factor.

\paragraph{Code retrieval}
\tabibert\ substantially outperforms Turkish models: 56.95 NDCG@10 versus \berturk\ (54.54, +2.41) and \ytubert\ (53.80, +3.15).
We attribute the improvement to exposure to code during pretraining and to the tokenizer's treatment of code syntax, and possibly to the longer context for long code snippets; we did not ablate these factors separately, so the gain is measurable but its attribution among them is not.

\subsubsection{Comparison with Multilingual mmBERT}
\label{sec:mmbert-comparison}

\mmbert\ achieves 78.98, 1.70 higher than \tabibert\ on the 0--100 average.
This reflects fundamental resource differences rather than architectural superiority, as both share the \modernbert\ architecture with different specialization strategies.

\paragraph{Resource differences}

\mmbert\ has 2.1 times as many parameters as \tabibert\ (307M versus 149M). Both are 22-layer ModernBERT-base models with approximately \num{110}M non-embedding parameters each. 
The difference lies almost entirely in the embedding matrix, which makes up about 38.5M of \tabibert's parameters (\num{50176} $\times$ 768) versus about 196.6M of \mmbert's (\num{256000} $\times$ 768). \mmbert's larger embedding matrix, needed to accommodate \num{1800} languages, accounts for its larger overall parameter count and higher memory requirements.

In tokenization, \mmbert's tokenizer shows 41 per cent higher fertility on Turkish text, requiring 41 per cent more tokens for identical content. This increases computational cost and reduces effective context by about \num{29} per cent (equivalent to a reduction from \num{8192} to approximately \num{5800} tokens of Turkish text). Finer tokenization may capture morphological nuances via character n-gram fragmentation but sacrifices efficiency and effective context length.

In training scale, \mmbert\ was pretrained on 3T tokens against \tabibert's 1T. Its massive multilingual corpus (\num{1800} languages) enables cross-lingual transfer, particularly from high-resource languages like English.

\paragraph{Practical trade-offs}
Despite \mmbert's higher scores, \tabibert\ offers advantages for Turkish-focused applications: fewer tokens per Turkish input and hence less computation per input (not a measured speedup), 
51 per cent fewer parameters (149M against 307M), and about 41 per cent longer effective Turkish context (\mmbert's is about 29 per cent shorter).
For Turkish-only deployment under memory or latency constraints, or for inputs longer than 512 tokens, \tabibert\ is the appropriate choice; where cross-lingual transfer or coverage of other languages is needed and the larger model is affordable, \mmbert\ scores higher; \berturk\ remains a strong choice for sentence-level classification, tagging and similarity on short inputs, where it is within 1 of \tabibert, but it cannot process inputs beyond 512 tokens, has no code or mathematics in its pretraining data, and trails by 9.55 F1 on question answering and 2.41 NDCG@10 on code retrieval.

\section{Discussion}
\label{sec:discussion}

Figure~\ref{fig:performance_comparison} visualizes performance patterns across \tabibench.
This section interprets these results, discusses architectural trade-offs, and addresses methodological considerations that informed our evaluation design.

\subsection{Interpreting Performance Patterns}

\paragraph{Task saturation versus architectural advantages}
On the four text-understanding categories (classification, token classification, STS, NLI) the leading Turkish models lie within 1 of one another, and these benchmarks approach saturation for base-sized encoders (Section~\ref{sec:results}).
\berturk's durability is itself informative: it has the smallest pretraining corpus of the baselines (35 GB against 75 GB for \ytubert\ and 110 GB for \tweetbert; Table~\ref{tab:results-summary}) yet outscores both, which points to corpus curation rather than data volume as the source of its strength.
Critically, \tabibert's competitive performance on these tasks demonstrates that long-context optimization does not compromise short-text understanding, a key validation of ModernBERT's architectural design.

The gains on question answering (+9.55 F1) and code retrieval (+2.41 NDCG@10) are where extended context, code in pretraining and the tokenizer provide measurable advantages, while the leads on academic understanding (+0.77 macro-F1) and information retrieval (+0.60 NDCG@10) are below 1 and we do not read them as decisive.
The context-length ablation in Appendix~\ref{sec:context-ablation} shows that this question answering gain is only partly a context-length effect; most of it persists even when \tabibert\ is capped at the same 512-token context as the baseline models.

\paragraph{Architectural innovations transfer to Turkish}
\tabibert's results indicate that \modernbert's innovations transfer to a morphologically rich, agglutinative language: the model matches or exceeds the Turkish baselines while inheriting the architecture's efficiency at long context and a sixteenfold longer context window than earlier BERT-based Turkish models~\citep{Warner2024ModernBERT}.

\subsection{Methodological Considerations and Design Decisions}

\paragraph{Tokenization-agnostic evaluation for token classification}
For named entity recognition and part-of-speech tagging, we adopted word-level rather than subword-level evaluation.
This addresses a fundamental comparability issue: different tokenizers fragment words into varying numbers of subword tokens.
A higher-fertility tokenizer would achieve artificially inflated subword-level scores by correctly predicting more fragmented tokens per word, even with identical word-level accuracy.

Word-level evaluation aggregates predictions at word boundaries before computing metrics, ensuring fair comparison regardless of tokenization strategy.
While subword-level metrics may be appropriate for analyzing tokenizer-specific behavior, cross-model comparisons require tokenization-agnostic metrics.
This principle is broadly relevant in multilingual NLP, where tokenization strategies vary substantially.

\paragraph{Pretraining infrastructure and reproducibility documentation}
During initial pretraining, we observed training throughput of about \num{30} per cent of that reported by \citet{Warner2024ModernBERT} on 8$\times$H100 SXM GPUs.
Experiments with 8$\times$A100 SXM showed only marginal decreases, suggesting data loading bottlenecks rather than GPU compute limitations.
We traced this with the \modernbert\ maintainers via a GitHub issue,\footnote{\texttt{\href{https://github.com/AnswerDotAI/ModernBERT/issues/236}{github.com/AnswerDotAI/ModernBERT/issues/236}}}
where we learned that the reported speeds relied on a pre-tokenized dataset, an implementation detail not documented in the paper or codebase; the maintainers have since pointed to data-preparation documentation in their follow-up Ettin repository~\citep{Weller2026SeqVsSeq}.
Implementing pre-tokenization resolved the throughput issue.

Implementation decisions that seem obvious to a model's developers must be stated explicitly if others are to reproduce their results; we document ours in Section~\ref{sec:tabibert}.

\section{Limitations}
\label{sec:limitations}

We acknowledge limitations constraining this work's scope and indicating areas for future research.

\subsection{Computational Resource Constraints}

\paragraph{Model scale}
We pretrained only base-sized models (149M parameters) due to budget limitations.
 \modernbert\ includes a large variant (395M parameters).
Scaling experiments with the \modernbert-large configuration could yield stronger performance, particularly on tasks requiring greater capacity (complex reasoning, low-resource specialized domains).

\paragraph{Corpus size and training duration}
Our pretraining corpus (86.58B tokens) is modest compared to large-scale corpora like FineWeb (15T tokens)~\citep{Penedo2024FineWeb}.
\tabibert\ was pretrained for 1T tokens versus \modernbert\ (2T) and \mmbert\ (3T).
Limited Turkish text availability necessitated \num{11.55} epochs over the same data, raising potential overfitting concerns.
However, we hypothesize that base model capacity (149M parameters) substantially mitigates memorization risks; that is, the model likely lacks sufficient capacity to memorize the corpus.
Larger, more diverse Turkish corpora would directly address this limitation.

\subsection{Evaluation Constraints}

\paragraph{Statistical significance}
Every score in this paper is a single fine-tuning run at the configuration selected on the validation split (Section~\ref{sec:evaluation-methodology}, seed 25), for each of the 27 datasets and five models.
A full multi-seed repetition was beyond our budget, since the published grid already amounts to on the order of 2,000 fine-tuning runs; we did, however, rerun the closest-margin categories with five seeds (Appendix~\ref{sec:multi-seed}): the NLI lead holds at +0.11 and the token-classification deficit narrows to $-0.06$, neither of which is significant against the seed spread, and the \tabibert-only reruns move the STS, information retrieval and academic understanding scores by at most 0.36.
Margins below 1 on the 0--100 scale, which is where all differences on the four short-text categories and three of \tabibert's five leads fall, should therefore be read as indicative rather than established; the question answering and code retrieval gains (9.55 F1 and 2.41 NDCG@10) are an order of magnitude larger than those margins.

\paragraph{Long-context evaluation gap}
\tabibert's \num{8192}-token window is sixteen times that of the earlier Turkish models, yet \tabibench\ gives limited opportunity to exercise it: most inputs are sentences or paragraphs (text and token classification, NLI, STS) or query-document pairs (retrieval).
The two datasets in which a material share of inputs exceeds 512 tokens are TQuAD (11--14 per cent of examples per split under the Turkish tokenizers) and Thesis-Abstract-Classification-11K (95th percentile 1,152 tokens); the context-length ablation in Appendix~\ref{sec:context-ablation} isolates the effect on the former.
Document-level QA, multi-document reasoning and legislative analysis are absent from Turkish evaluation infrastructure, a gap of the field rather than of this work, but one that constrains validation of the long-context capability.

\subsection{Translation-Based Datasets: Rationale and Limitations}
\label{sec:translation-limitations}
Fifteen of the twenty-seven \tabibench\ datasets are machine-translated and two are human-translated (Table~\ref{tab:tabibench-datasets} gives the translating agent and source of each): six were translated by us using GPT-4.1 (the four code retrieval and two academic understanding datasets); four information retrieval datasets by the TR-MTEB group using \texttt{aya-expanse-8b}~\citep{BaysanGungor2025TR-MTEB} and MsMarco-TR by prior work~\citep{Kesgin2023YTU}; and SNLI-TR, MultiNLI-TR~\citep{BudurEtAl2020Data}, STSb-TR~\citep{BekenFikri2021Summarization}, and SICK-TR~\citep{Dehghan2025Turkish} by their original authors; XQuAD~\citep{Artetxe2019CrossLingual} was translated professionally as part of its original construction, and BiText combines professionally translated WMT16 news text with the SETIMES2 parallel corpus~\citep{Bojar2016WMT16}.
The translation systems behind these resources differ considerably: NLI-TR was produced with Amazon Translate and validated by bilingual experts~\citep{BudurEtAl2020Data}, STSb-TR and SICK-TR with the Google Cloud Translation API, the former checked by BLEU against reference translations and the latter unverified~\citep{BekenFikri2021Summarization,Dehghan2025Turkish}, the four TR-MTEB sets with the 8B-parameter aya-expanse model~\citep{BaysanGungor2025TR-MTEB}, and MsMarco-TR with a system that is not documented; for our own six datasets we used GPT-4.1, a frontier general-purpose model substantially larger and more recent than any of these, and we validated the result as described below.
We opted to rely on translations from a pragmatic stance.
High-quality Turkish datasets for retrieval and specialized domains are scarce or non-existent, 
yet these tasks are critical for comprehensive encoder evaluation.
For code retrieval, only natural language components are translated while code remains unchanged.
To substantiate the quality of the six translations we produced ourselves rather than assert it, we ran a native-speaker adequacy audit.
Three native Turkish speakers, graduate and undergraduate students of computer engineering at Boğaziçi University who are not authors of this paper, independently rated a stratified random sample of \num{120} test-split items (20 per dataset) against the English source on a three-level scale (meaning preserved; meaning preserved but flawed; meaning changed) and flagged whether the gold label or gold document still held for the Turkish version; the protocol is given in Appendix~\ref{sec:audit-protocol}.
By majority vote, the translation preserved the meaning of the source in 108 of the 120 items (90.0 per cent; 95 per cent Wilson interval 83.3--94.2) and the gold label or gold document still held for 119 (99.2 per cent; 95.4--99.9): 92 items (76.7 per cent) were rated fully adequate, 16 (13.3 per cent) preserved meaning with lesser flaws such as awkward phrasing or a non-standard term, and 12 (10.0 per cent; 5.8--16.7) contained a meaning-changing error, between one and three per dataset (three-way splits, three items, were resolved to the more severe label).
Agreement on whether meaning was preserved was moderate (Fleiss' $\kappa = 0.51$; 0.24 on the three-level scale, where one rater applied a markedly more lenient threshold for lesser flaws).
The meaning-changing errors were mostly omitted titles or sections in long inputs, altered numeric masks in PubMedRCT-10K-TR and domain-terminology slips, and they concentrate in outputs of unusual length: a separately sampled stratum of 17 length-ratio outliers had a 65 per cent error rate, so a simple length-ratio check identifies most gross failures.
We report these figures rather than re-translate the flagged items, so that the released datasets remain the ones evaluated here, and we release the audit scripts and anonymized ratings with the benchmark.

Machine-translated content may fail to capture linguistic nuance, idiomatic expressions or cultural context.
However, in our case, all of the content is technical or formal in nature, minimizing such risks.
Moreover, omitting such task categories with the aim of avoiding machine-translated content would severely limit the evaluation scope.

\section{Conclusions and Future Work}
\label{sec:conclusions}

This work introduces two resources for Turkish NLP: \tabibert, a ModernBERT-based encoder demonstrating that recent architectural advances transfer effectively to Turkish; and \tabibench, a standardized evaluation benchmark addressing reproducibility challenges in Turkish NLP research.

\subsection{Key Contributions}

\paragraph{TabiBERT}
We developed a Turkish encoder on the \modernbert\ architecture, pretrained for 1T tokens on an 86.58B-token corpus of web, scientific, parliamentary and academic text supplemented with bilingual, code and mathematical data, with a Turkish-specialized tokenizer.
\tabibert\ achieves an average of 77.28 on \tabibench, leading the Turkish models in five of eight task categories and \berturk\ in six of eight, with its largest advantages on question answering (+9.55 F1) and code retrieval (+2.41 NDCG@10).
It offers an \num{8192}-token context, sixteen times that of earlier Turkish models, with 51 per cent fewer parameters than \mmbert\ (149M against 307M) and, because \mmbert\ needs 41 per cent more tokens for Turkish, about 41 per cent longer effective Turkish context; ModernBERT-based encoders also offer faster inference at long context lengths \citep{Warner2024ModernBERT}.
While \mmbert\ scores higher (78.98), this gap of 1.70 reflects 2.1$\times$ parameters and 3$\times$ training tokens, not architectural differences.

\paragraph{\tabibench}
We developed a standardized benchmark of 27 curated datasets across eight task categories with fixed splits, task-appropriate metrics and a fair comparison protocol under which every model is fine-tuned identically, and we release it as an open resource and as a template for similar efforts in other lower-resource languages.

\paragraph{Methodological insights}
Documenting the pretraining infrastructure challenges we encountered highlights the importance of reporting implementation details for reproducibility.
Adopting tokenization-agnostic metrics (word-level evaluation for token classification) addresses a comparability issue relevant across multilingual NLP.

\subsection{Broader Implications}

Beyond Turkish, the methodology (systematic dataset quality review, language-specific optimization and transparent reporting of constraints) generalizes to other languages that lack standardized evaluation frameworks, and the results support targeted, resource-efficient development as a complement to large-scale multilingual approaches, the choice between them depending on application requirements rather than on universal superiority.

\subsection{Future Directions}

The most significant gap is the absence of Turkish datasets that evaluate long-context understanding (long and multi-document question answering, legislative analysis); developing them would allow the \num{8192}-token context to be assessed directly.
Comprehensive multi-seed significance testing and robustness analysis under input perturbations (typos, morphological variation, domain shift) would formalize the comparisons reported here.
Pretraining a larger variant (\modernbert-large, 395M parameters) and training beyond 1T tokens on expanded corpora would show whether the advantages scale with capacity and data.
Continued training from \tabibert\ optimized for retrieval could provide Turkish NLP with a specialized embedding model potentially superior to multilingual alternatives, and continued pretraining for specialized domains (legal, medical, scientific) could reveal domain-specific benefits.
A controlled comparison of Turkish-specific and multilingual \modernbert\ variants at matched compute and corpus quality would quantify the benefits of monolingual specialization against cross-lingual transfer.
Finally, \tabibench\ can expand to further task types, such as mathematical understanding, and be maintained as a living benchmark with versioned releases.

\tabibert\ and \tabibench\ contribute to Turkish NLP by providing both an efficient, modern encoder model and a standardized evaluation benchmark.
These resources address infrastructure gaps that have hindered progress, and we believe they will facilitate more rigorous and reproducible research in Turkish language technology.
\section*{Acknowledgments}
We thank Mehmet Selman Baysan for his contributions to the implementation of the retrieval tasks, and Mehmet Selman Baysan, Hüseyin Emir Akdağ and Muhammed Emin Arayıcı, PhD, MSc and BSc students in the Computer Engineering Department of Boğaziçi University, respectively, for rating the translation adequacy audit sample.
Claude Fable 5.1 and Claude Opus 5.5 (Anthropic) were used for grammar checking, language refinement, and run scripted consistency checks on the reported numbers and cross-references. Every suggestion was reviewed and edited by the authors, no experimental result or analysis was produced by AI, and the authors take full responsibility for the content.

\section*{Funding Statement}
Computational resources for pretraining and fine-tuning were funded by VNGRS\footnote{\href{https://vngrs.com/}{https://vngrs.com/}}; additional GPU resources were provided by Lambda Labs\footnote{\href{https://lambda.ai/}{https://lambda.ai/}}.
The authors received no other funding for this work.

\section*{Competing Interests}
The authors declare the following financial interests/personal relationships which may be considered as potential competing interests:
Melikşah Türker is a PhD student at Boğaziçi University and is employed by VNGRS, which provided the cloud GPU infrastructure for model training.
Dr. Onur Güngör is a part-time faculty member at Boğaziçi University and is also employed by Udemy.
While these entities may promote this research for institutional branding or professional visibility, the authors received no direct financial benefits from these activities.
All training, evaluation, and manuscript preparation were conducted independently by the research lab to ensure transparency and reproducibility.
The companies had no role in the study design or the reporting of results.
In addition to these financial interests, \tabibench\ was curated, standardized, and released by the same authors who developed \tabibert\ and who use \tabibench\ to evaluate it in this paper;
Section~\ref{sec:tabibench-governance} documents the safeguards that keep the benchmark's construction independent of this evaluation.
This authorship overlap applies to all authors equally, not only those named above; beyond it and the financial interests already stated, the authors declare no other competing interests.

\section*{Ethics Statement}
Ethical approval was not required for this study.
It involved no human participants or animal subjects and collected no personal data; it used existing text corpora and datasets, and the translation-quality audit was carried out by volunteer annotators who rated published text and are acknowledged by name with their consent.

\section*{Data Availability Statement}
\tabibert's model weights are available on Hugging Face~\citep{TabiBERT-HF}, and its pretraining and fine-tuning code, training configurations and the per-seed scores of Appendix~\ref{sec:multi-seed} on GitHub~\citep{TabiBERT}.
\tabibench\ is released on Hugging Face~\citep{TabiBench} as a collection of 27 datasets, each redistributed under the terms of its original source as recorded in Table~\ref{tab:benchmark_governance}; MedNLI-TR is excluded for the reasons given in Section~\ref{sec:tabibench-governance}.
The translation audit scripts and anonymized ratings are released with the code.
The pretraining corpus is not redistributed, because portions of it (academic publications and books from institutional repositories) are subject to licensing restrictions; its sources, sizes and sampling rates are given in Section~\ref{sec:tabibert} and Table~\ref{tab:train_data}, and comparable corpora can be assembled from the publicly available sources listed there.

\bibliography{references}

\appendix

\section{Fine-Tuning Results}
\label{sec:appendix}

\subsection{Per-Dataset Fine-Tuning Results}
The appendix presents detailed performance comparisons across all evaluation tasks.
Each table compares \tabibert\ against established monolingual Turkish models, with the best-performing monolingual model highlighted in each task.
We also include \mmbert\ as a reference point.
While the very large-scale multilingual model \mmbert\ often achieves competitive or superior performance, direct comparison is complicated by substantial differences in model scale, training data volume, and cross-lingual training objectives.

\subsubsection{Text Classification}
Model performance is evaluated across four text classification datasets:
NewsCat \citep{McemilgNewsCat},
BilTweetNews \citep{Toraman2017EventPrediction,Toraman2021EventRetrieval},
GenderHateSpeech \citep{Toraman2022HateSpeech,Sahinuc2023GenderBias}, and
ProductReviews \citep{FthbrmnbyTurkishProductReviews}.
Table~\ref{tab:classification} presents these results using the macro-F1 score.
Notably, \mmbert\ (82.54) ranks below three of the four Turkish models, exceeding only \tweetbert\ (79.71).

\begin{table}[!ht]
 \centering
    \caption{Downstream performance of models on Text classification (eval metric: macro-F1).
    The number of test samples per task is also given. For each column, the highest score among the Turkish models (excluding \mmbert) is shown in bold.}
    \label{tab:classification}
    \resizebox{\linewidth}{!}{
    \begin{tabular}{@{}lSSSSS@{}}
    \toprule
    \textbf{Model} & \textbf{NewsCat} & \textbf{BilTweetNews} & \textbf{GenderHateSpeech} & \textbf{ProductReviews} & \textbf{Weighted avg.} \\
    \textit{Test size} & \multicolumn{1}{c}{\num{250}} & \multicolumn{1}{c}{\num{150}} & \multicolumn{1}{c}{\num{2000}} & \multicolumn{1}{c}{\num{35275}}& \\ \midrule
    \tweetbert\ & 91.98 & 53.16 & 68.58 & 80.37 & 79.71 \\
    \ytubert\ & \B 97.21 & 53.39 & \B 71.02 & \B 85.04 & \B 84.25 \\
    \berturk\ & 95.60 & \B 57.87 & 68.25 & 84.30 & 83.42 \\
    \tabibert\ & 95.20 & 50.11 & 69.01 & 84.32 & 83.44 \\ \midrule
    \mmbert\ & 94.80 & 49.06 & 66.45 & 83.51 & 82.54 \\ \bottomrule
\end{tabular}}
\end{table}

\subsubsection{Token Classification}
Token classification performance is evaluated using four datasets:
WikiNER \citep{Altinok2023Turkish},
WikiANN-TR \citep{Rahimi2019WikiANN},
PosUD-BOUN \citep{UDTurkishBOUN},
PosUD-IMST \citep{UDTurkishIMST}.
Table~\ref{tab:token_classification} presents the micro-F1 scores.
The \berturk\ model achieves the highest overall score of \num{93.67}, leading two of the four datasets (WikiNER and PosUD-IMST),
while \ytubert\ leads WikiANN-TR (95.41), and \tabibert\ leads PosUD-BOUN (90.40).

\begin{table}[!ht]
   \centering
      \caption{Downstream performance of models on Token classification (eval metric: micro-F1). The number of test samples per task is also given. For each column, the highest score among the Turkish models (excluding \mmbert) is shown in bold.}
      \label{tab:token_classification}
      \resizebox{0.8\linewidth}{!}{
      \begin{tabular}{@{}lSSSSS@{}}
      \toprule
      \textbf{Model} & \textbf{WikiNER} & \textbf{WikiANN-TR} & \textbf{PosUD-BOUN} & \textbf{PosUD-IMST} & \textbf{Weighted avg.} \\
      \textit{Test size} & \multicolumn{1}{c}{\num{1000}} & \multicolumn{1}{c}{\num{10000}} & \multicolumn{1}{c}{\num{979}} & \multicolumn{1}{c}{\num{1100}} & \\ \midrule
      \tweetbert\ & 72.6 & 94.05 & 89.74 & 93.21 & 92.02 \\
      \ytubert\ & 78.96 & \B 95.41 & 89.19 & 94.30 & 93.60 \\
      \berturk\ & \B 79.44 & 95.37 & 89.82 & \B 94.60 & \B 93.67 \\
      \tabibert\ & 76.41 & 95.34 & \B 90.40 & 94.07 & 93.42 \\ \midrule
      \mmbert\ & 75.97 & 95.85 & 90.72 & 94.26 & 93.81 \\ \bottomrule
  \end{tabular}}
\end{table}

\subsubsection{Semantic Textual Similarity}
The performance in semantic textual similarity is evaluated with two datasets:
SICK-TR \citep{Dehghan2025Turkish} and
STSb-TR \citep{BekenFikri2021Summarization}.
Table~\ref{tab:semantic_similarity} reports the results using the Pearson correlation score.
All models, with the exception of \tweetbert, achieve similar results with \berturk\ performing the best with \num{85.33}.

\begin{table}[!ht]
    \centering
       \caption{Downstream performance of models on Semantic textual similarity (eval metric: Pearson correlation). The number of test samples per task is also given. For each column, the highest score among the Turkish models (excluding \mmbert) is shown in bold.}
       \label{tab:semantic_similarity}
       \small
       \resizebox{0.5\linewidth}{!}{
       \begin{tabular}{@{}lSSS@{}}
       \toprule
       \textbf{Model} & \textbf{SICK-TR} & \textbf{STSb-TR} & \textbf{Weighted avg.} \\
       \textit{Test size} & \multicolumn{1}{c}{\num{4927}} & \multicolumn{1}{c}{\num{1379}} & \\ \midrule
       \tweetbert\ & 78.35 & 66.96 & 75.86 \\
       \ytubert\ & 85.27 & 82.55 & 84.68 \\
       \berturk\ & \B 85.95 & 83.12 & \B 85.33 \\
       \tabibert\ & 85.00 & \B 83.84 & 84.74 \\ \midrule
       \mmbert\ & 87.81 & 84.31 & 87.05 \\ \bottomrule
   \end{tabular}}
\end{table}

\subsubsection{Natural Language Inference}

Natural language inference capabilities are evaluated using two Turkish datasets:
SNLI-TR and
MultiNLI-TR \citep{BudurEtAl2020Data}.
Table~\ref{tab:nli_results} reports model performance based on the macro-F1 score.
With the exception of \tweetbert, the models exhibit similar performance levels with macro-F1 scores near \num{84}.

\begin{table}[!ht]
    \centering
       \caption{Downstream performance of models on Natural language inference (eval metric: macro-F1). The number of test samples per task is also given. For each column, the highest score among the Turkish models (excluding \mmbert) is shown in bold.}
       \label{tab:nli_results}
       \small
       \resizebox{0.5\linewidth}{!}{
       \begin{tabular}{@{}lSSS@{}}
       \toprule
       \textbf{Model} & \textbf{SNLI-TR} & \textbf{MultiNLI-TR} & \textbf{Weighted avg.} \\
       \textit{Test size} & \multicolumn{1}{c}{\num{9824}} & \multicolumn{1}{c}{\num{4923}} & \\ \midrule
       \tweetbert\ & 83.05 & 71.21 & 79.10 \\
       \ytubert\ & 87.18 & 78.15 & 84.16 \\
       \berturk\ & \B 87.21 & 78.57 & 84.33 \\
       \tabibert\ & 86.47 & \B 80.60 & \B 84.51 \\ \midrule
       \mmbert\ & 86.44 & 80.28 & 84.38 \\ \bottomrule
   \end{tabular}}
\end{table}

\subsubsection{Question Answering}
Question answering performance is evaluated using two Turkish datasets:
TQuAD~\citep{ErdometoTQuad2} and
XQuAD~\citep{Artetxe2019CrossLingual}.
Table~\ref{tab:qa_results} presents the F1 scores obtained by each model.
Among all evaluated tasks, this benchmark reveals the most substantial performance gain achieved by \tabibert\ over other Turkish models with an F1 score of \num{69.71} in comparison to the next best result of \num{60.16} recorded by \berturk.

Two factors contribute: the model itself, trained on our corpus with our tokenizer, and the context cap, since \tabibert\ and \mmbert\ were fine-tuned on TQuAD with a \num{1024}-token cap, under which about 95 per cent of its examples are untruncated, whereas the other Turkish models are limited to \num{512} tokens (all models used \num{384} tokens for XQuAD); Appendix~\ref{sec:context-ablation} separates the two and attributes about 2 of the 9.04 F1 gap on TQuAD to context length.

As for XQuAD, its small training split (\num{833} examples) makes single-seed fine-tuning unstable, which explains the wide spread of Turkish-model scores better than data scarcity alone, since \mmbert\ reaches \num{70.40} on the identical split.

\begin{table}[!ht]
    \centering
       \caption{Downstream performance of models on Question answering (eval metric: F1). The number of test samples per task is also given. For each column, the highest score among the Turkish models (excluding \mmbert) is shown in bold.}
       \label{tab:qa_results}
       \small
       \resizebox{0.5\linewidth}{!}{
       \begin{tabular}{@{}lSSS@{}}
       \toprule
       \textbf{Model} & \textbf{TQuAD} & \textbf{XQuAD} & \textbf{Weighted avg.} \\
       \textit{Test size} & \multicolumn{1}{c}{\num{2520}} & \multicolumn{1}{c}{\num{179}} & \\ \midrule
       \tweetbert\ & 38.04 & \B 39.40 & 38.13 \\
       \ytubert\ & 32.01 & 24.25 & 31.50 \\
       \berturk\ & 63.30 & 15.96 & 60.16 \\
       \tabibert\ & \B 72.34 & 32.61 & \B 69.71 \\ \midrule
       \mmbert\ & 71.55 & 70.40 & 71.47 \\ \bottomrule
   \end{tabular}}
\end{table}

\subsubsection{Information Retrieval}
We evaluate information retrieval on six datasets (Table~\ref{tab:tabibench-datasets}): BiText, human-translated WMT16/SETIMES2 parallel text~\citep{Bojar2016WMT16}; MsMarco-TR, a pre-existing machine translation attributed to \citet{Kesgin2023YTU}; and Scifact-TR, Fiqa-TR, NFCorpus-TR and Quora-TR, translated by the TR-MTEB group~\citep{BaysanGungor2025TR-MTEB}.
Table~\ref{tab:retrieval_results} reports the NDCG@10 scores achieved by each model.
Across these six retrieval benchmarks, \tabibert\ achieves the highest weighted-average NDCG@10 among Turkish models (75.44),
leading on BiText and MsMarco-TR, the two largest sets, while trailing \ytubert\ on Scifact-TR, Fiqa-TR and Quora-TR and \berturk\ on NFCorpus-TR.
We attribute these gains in part to \tabibert's exposure to more diverse pretraining data, including bilingual sources, which appear to be particularly beneficial for retrieval-oriented tasks.
Its largest deficit, 5.66 NDCG@10 behind \ytubert\ on Scifact-TR, falls on the smallest of the six datasets (\num{339} test instances).

\begin{table}[!ht]
    \centering
       \caption{Downstream performance of models on Retrieval tasks (eval metric: NDCG@10). The number of test samples per task is also given. For each column, the highest score among the Turkish models (excluding \mmbert) is shown in bold.}
       \label{tab:retrieval_results}
       \small
       \resizebox{\linewidth}{!}{
       \begin{tabular}{@{}lSSSSSSS@{}}
       \toprule
       \textbf{Model} & \textbf{BiText} & \textbf{MsMarco-TR} & \textbf{Scifact-TR} & \textbf{Fiqa-TR} & \textbf{NFCorpus-TR} & \textbf{Quora-TR} & \textbf{Weighted avg.} \\
       \textit{Test size} & \multicolumn{1}{c}{\num{3000}} & \multicolumn{1}{c}{\num{31692}} & \multicolumn{1}{c}{\num{339}} & \multicolumn{1}{c}{\num{1706}} & \multicolumn{1}{c}{\num{12334}} & \multicolumn{1}{c}{\num{15675}} & \\ \midrule
       \tweetbert\ & 95.98 & 74.20 & 68.25 & 36.23 & 27.79 & 86.84 & 68.40 \\
       \ytubert\ & 97.25 & 81.99 & \B 79.88 & \B 51.10 & 28.36 & \B 92.87 & 74.29 \\
       \berturk\ & 96.77 & 81.73 & 78.47 & 49.46 & \B 32.47 & 92.72 & 74.84 \\
       \tabibert\ & \B 99.42 & \B 83.31 & 74.22 & 50.67 & 31.19 & 92.47 & \B 75.44 \\ \midrule
       \mmbert\ & 99.49 & 83.68 & 80.00 & 52.29 & 33.44 & 92.78 & 76.20 \\ \bottomrule
   \end{tabular}}
\end{table}

\subsubsection{Code Retrieval}
We evaluated code retrieval performance using four datasets: 
Apps-TR, CosQA-TR, StackOverflowQA-TR, CodeSearchNet-21K-TR~\citep{TabiBERTCodeRetrievalDatasets}.
Table~\ref{tab:code_retrieval_results} presents the NDCG@10 scores obtained by each model.
In code retrieval tasks, \tabibert\ outperforms all other Turkish models, achieving an average NDCG@10 score of \num{56.95}, considerably higher than the next best result of \num{54.54} by \berturk.
We attribute these strengths to the incorporation of code in pretraining data and to the tokenizer's treatment of code syntax, and possibly to its support for longer contexts; these factors were not ablated separately (Section~\ref{sec:results}).
Across the individual tasks, \tabibert\ secures the top NDCG@10 score in three out of four datasets, 
and is second on Apps-TR, trailing \berturk\ by 0.69 NDCG@10 (17.49 against 18.18).

\begin{table}[!ht]
    \centering
       \caption{Downstream performance of models on Code retrieval tasks (eval metric: NDCG@10). The number of test samples per task is also given. For each column, the highest score among the Turkish models (excluding \mmbert) is shown in bold.}
       \label{tab:code_retrieval_results}
       \small
       \resizebox{\linewidth}{!}{
       \begin{tabular}{@{}lSSSSS@{}}
       \toprule
       \textbf{Model} & \textbf{Apps-TR} & \textbf{CosQA-TR} & \textbf{StackOverflowQA-TR} & \textbf{CodeSearchNet-21K-TR} & \textbf{Weighted avg.} \\
       \textit{Test size} & \multicolumn{1}{c}{\num{3770}} & \multicolumn{1}{c}{\num{500}} & \multicolumn{1}{c}{\num{1994}} & \multicolumn{1}{c}{\num{3000}} & \\ \midrule
       \tweetbert\ & 04.48 & 82.32 & 67.01 & 70.41 & 43.49 \\
       \ytubert\ & 14.47 & 82.04 & 82.65 & 79.35 & 53.80 \\
       \berturk\ & \B 18.18 & 85.72 & 79.93 & 78.15 & 54.54 \\
       \tabibert\ & 17.49 & \B 88.11 & \B 82.85 & \B 84.12 & \B 56.95 \\ \midrule
       \mmbert\ & 32.24 & 89.31 & 90.43 & 88.36 & 66.02 \\ \bottomrule
   \end{tabular}}
\end{table}

\subsubsection{Academic Understanding Tasks}
We evaluated academic understanding tasks using three datasets: 
PubMedRCT-10K-TR, SciCite-TR, Thesis-Abstract-Classification-11K~\citep{TabiBERTAcademicDomainDatasets}.
All three are text classification datasets.
Table~\ref{tab:academic_results} presents the results obtained by each model, reporting the macro-F1 scores.
\tabibert\ leads this category among Turkish models by a small margin (+0.77 macro-F1 over \ytubert).
On the demanding task of Thesis-Abstract-Classification-11K, characterized by long input contexts (with a 95th percentile context length of \num{1152} tokens) and a sparse number of samples for each class (a sum of 60 examples per class throughout all splits), \tabibert\ attains the highest score among Turkish models with \num{50.77}, outperforming the next best model \berturk\ by 1.57 macro-F1.
We attribute this result to a pretraining corpus rich in Turkish scientific articles and graduate theses and to the ability to process longer inputs.

\begin{table}[!ht]
    \centering
       \caption{Downstream performance of models on Academic understanding tasks (eval metric: macro-F1). The number of test samples per task is also given. For each column, the highest score among the Turkish models (excluding \mmbert) is shown in bold.}
       \label{tab:academic_results}
       \small
       \resizebox{\linewidth}{!}{
       \begin{tabular}{@{}lSSSS@{}}
       \toprule
       \textbf{Model} & \textbf{PubMedRCT-10K-TR} & \textbf{SciCite-TR} & \textbf{Thesis-Abstract-Classification-11K} & \textbf{Weighted avg.} \\
       \textit{Test size} & \multicolumn{1}{c}{\num{1500}} & \multicolumn{1}{c}{\num{1859}} & \multicolumn{1}{c}{\num{1683}} & \\ \midrule
       \tweetbert\ & 70.07 & 79.59 & 31.18 & 60.60 \\
       \ytubert\ & 75.32 & \B 84.09 & 47.56 & 69.29 \\
       \berturk\ & \B 75.61 & 81.60 & 49.20 & 69.00 \\
       \tabibert\ & 75.32 & 83.29 & \B 50.77 & \B 70.06 \\ \midrule
       \mmbert\ & 74.37 & 83.27 & 52.47 & 70.34 \\ \bottomrule
   \end{tabular}}
\end{table}

\subsection{Context-Length Ablation on Question Answering}
\label{sec:context-ablation}

Among all benchmark tasks, question answering shows the largest gap between \tabibert\ and the next-best Turkish model: 
on TQuAD specifically, \tabibert\ scores \num{72.34} against \berturk's \num{63.30}, a difference of 9.04 F1.
Since \tabibert\ and \mmbert\ are the only models fine-tuned on TQuAD with a \num{1024}-token cap rather than \num{512} (Section~\ref{sec:evaluation-methodology}), 
part of this gap could simply reflect their access to a longer context rather than a genuine modeling advantage.
This section decomposes the gap into a context-length component and a model component.

We first measure how much of the input is actually affected by a \num{512}-token cap.
Since all models were fine-tuned on XQuAD with the same \num{384}-token cap (Section~\ref{sec:evaluation}) and only \num{1.35} per cent of its examples exceed \num{512} tokens, XQuAD offers no context-length differential between models and is therefore excluded from this analysis.
We instead focus on TQuAD, where \tabibert\ and \mmbert\ were fine-tuned with the longer cap and thus could plausibly benefit from it.
For each of the five models' tokenizers, we compute the fraction of TQuAD examples whose question and context together,
exactly as fed to the model when only the context is truncated (the question is always kept intact), exceed \num{512} tokens.
Table~\ref{tab:qa_length_over512} reports these shares per split.
\num{10.83}--\num{13.98} per cent of examples exceed \num{512} tokens for the four Turkish-oriented tokenizers,
with \mmbert's multilingual tokenizer pushing \num{14.96}--\num{25.35} per cent of examples over the limit due to higher subword fertility on Turkish text.

We therefore isolate the context-length effect on TQuAD, where it is material.
We fine-tune \tabibert\ and \mmbert\ again on TQuAD with the maximum context length capped at \num{512} tokens, 
the limit supported by \berturk, \tweetbert, and \ytubert, whose published scores are already obtained under this cap and require no new run.
For this ablation, we reuse each model's original best hyperparameter configuration selected during the main hyperparameter search (Section~\ref{sec:evaluation}) 
rather than repeating hyperparameter tuning at the new context length, since running a full second search for this ablation alone was not computationally feasible.
Table~\ref{tab:qa_results_512cap} reports the resulting scores alongside the originally published, uncapped ones.

Capping the context at \num{512} tokens lowers \tabibert's TQuAD F1 from \num{72.34} to \num{70.31} (\num{-2.03} F1) 
and \mmbert's from \num{71.55} to \num{69.26} (\num{-2.29} F1), confirming that some of both models' advantage does come from their longer supported context.
Crucially, however, the \num{512}-capped \tabibert\ still scores \num{70.31}, well above \berturk's \num{63.30}.
Decomposing the original 9.04 F1 TQuAD gap between \tabibert\ and \berturk\ accordingly, 
about 2 F1 (\num{72.34}\,$\to$\,\num{70.31}) are attributable to context length, 
and the remaining roughly 7 F1 (\num{70.31}\,$\to$\,\num{63.30}) reflect a genuine modeling advantage independent of context.
The same pattern holds for \mmbert\ (\num{512}-capped score \num{69.26}, still 5.96 F1 above \berturk), 
confirming that the higher performance of \tabibert\ and \mmbert\ on question answering is only partially explained by context length; 
most of the gap is a model effect.

This context effect is specific to long-passage datasets such as TQuAD and does not arise on XQuAD, whose \num{384}-token cap already accommodates essentially all examples for every model.

\begin{table}[!ht]
    \centering
       \caption{Share of TQuAD question answering examples exceeding 512 tokens under each model's tokenizer, by dataset split. Input length is measured as question + context, special tokens included, exactly as fed to the model (\texttt{truncation='only\_second'}); uncased models are lowercased first. Values are per cent of examples in each split.}
       \label{tab:qa_length_over512}
       \small
       \resizebox{0.5\linewidth}{!}{
       \begin{tabular}{@{}lSSS@{}}
       \toprule
       & \multicolumn{3}{c}{\textbf{TQuAD}} \\
       \cmidrule(lr){2-4}
       \textbf{Model} & \textbf{Train} & \textbf{Val} & \textbf{Test} \\
       \textit{Split Size} & \multicolumn{1}{c}{\num{11803}} & \multicolumn{1}{c}{\num{2418}} & \multicolumn{1}{c}{\num{2520}} \\ \midrule
       \berturk\ & 11.98 & 13.15 & 10.87 \\
       \tweetbert\ & 11.07 & 12.12 & 10.95 \\
       \ytubert\ & 11.32 & 11.95 & 10.83 \\
       \tabibert\ & 13.17 & 13.98 & 10.83 \\ \midrule
       \mmbert\ & 24.32 & 25.35 & 14.96 \\ \bottomrule
   \end{tabular}}
\end{table}

\begin{table}[!ht]
    \centering
       \caption{Published TQuAD question answering F1 scores versus scores obtained when the context is capped at \num{512} tokens (eval metric: F1).
       Published scores use \num{1024} tokens for \tabibert\ and \mmbert\ and \num{512} tokens, the maximum length they support, for \berturk, \tweetbert, and \ytubert.
       Since the three baseline models' TQuAD scores are already \num{512}-capped, only \tabibert\ and \mmbert\ required a new \num{512}-token TQuAD run; no other cell required a new fine-tuning run.
       These new runs reuse each model's original best hyperparameter configuration rather than repeating hyperparameter tuning, due to compute limitations.}
       \label{tab:qa_results_512cap}
       \small
       \resizebox{0.5\linewidth}{!}{
       \begin{tabular}{@{}lSSS@{}}
       \toprule
       & \multicolumn{3}{c}{\textbf{TQuAD}} \\
       \cmidrule(lr){2-4}
       \textbf{Model} & \textbf{Published} & \textbf{512-cap} & \textbf{$\Delta$} \\ \midrule
       \tweetbert\ & 38.04 & 38.04 & 0.00 \\
       \ytubert\ & 32.01 & 32.01 & 0.00 \\
       \berturk\ & 63.30 & 63.30 & 0.00 \\
       \tabibert\ & 72.34 & 70.31 & -2.03 \\ \midrule
       \mmbert\ & 71.55 & 69.26 & -2.29 \\ \bottomrule
   \end{tabular}}
\end{table}

\subsection{Multi-Seed Check of the Closest Margins}
\label{sec:multi-seed}
Every score in the main tables is a single fine-tuning run (seed 25).
To check how robust the smallest margins between \tabibert\ and \berturk\ are, we repeated fine-tuning and evaluation with five seeds (7, 25, 42, 100 and 1234) at each model's selected hyperparameter configuration: for both models on NLI and token classification, the categories in which \tabibert\ leads (+0.18) and trails ($-0.25$) by the smallest margins, and, due to compute limitations, for \tabibert\ only on the three further categories whose single-seed margins are below 1, STS ($-0.59$), information retrieval (+0.60) and academic understanding (+0.77).
On NLI, \tabibert\ scores $84.45 \pm 0.13$ (mean $\pm$ standard deviation across seeds) against $84.34 \pm 0.14$ for \berturk: the lead holds at +0.11 but is of the same size as the seed spread, and a Welch $t$-test on the five-seed category scores does not reach significance ($t = 1.23$).
On token classification, \tabibert\ scores $93.31 \pm 0.08$ against $93.37 \pm 0.58$ for \berturk: the deficit narrows from $-0.25$ to $-0.06$ and is likewise not significant ($t = -0.21$).
The \tabibert-only reruns give $85.10 \pm 0.44$ on STS, $75.61 \pm 0.17$ on information retrieval and $69.88 \pm 0.21$ on academic understanding, moving the single-seed scores by between 0.17 and 0.36 and leaving the direction of every margin unchanged: against the published \berturk\ scores, \tabibert\ still leads information retrieval (+0.77) and academic understanding (+0.59) and still trails STS ($-0.23$).
Seed variation is therefore of the same order as these margins, which is why Section~\ref{sec:limitations} treats margins below 1 as indicative rather than established; the per-seed scores are released with the code.

\subsection{TabiBench Source Inventory and Licensing}
\label{sec:appendix-governance}
Table~\ref{tab:benchmark_governance} accompanies Section~\ref{sec:tabibench-governance} and reports, for each of the 27 datasets that constitute \tabibench, its original provider,
the license or terms of use we were able to locate, the resulting redistribution status of that dataset within the public \tabibench\ release, and the processing we applied to it.

\begingroup
\footnotesize
\setlength{\tabcolsep}{4pt}
\renewcommand{\arraystretch}{1.15}
\begin{longtable}{p{2.1cm}p{3.4cm}p{2.9cm}p{1.2cm}p{2.9cm}}
    \caption{\tabibench\ source inventory: original provider (with a direct link to the source repository, where one exists), located license/terms, resulting redistribution status, and processing applied by the \tabibench\ authors, for each constituent dataset (Section~\ref{sec:tabibench-governance}).
    License information was compiled by consulting original dataset repositories, dataset cards, and associated publications where available; entries marked \textit{not declared} indicate that no license could be located, not that a dataset is confirmed to be unrestricted.
    This table reflects a best-effort review rather than an independent legal audit; readers with license-sensitive use cases should verify terms directly against the cited original sources.}
    \label{tab:benchmark_governance} \\
    \toprule
    \textbf{Dataset} & \textbf{Provider} & \textbf{License / terms} & \textbf{Redistrib.} & \textbf{Processing} \\
    \midrule
    \endfirsthead
    \multicolumn{5}{l}{\textit{Table~\ref{tab:benchmark_governance} continued}} \\
    \toprule
    \textbf{Dataset} & \textbf{Provider} & \textbf{License / terms} & \textbf{Redistrib.} & \textbf{Processing} \\
    \midrule
    \endhead
    \midrule
    \multicolumn{5}{r}{\textit{continued on next page}} \\
    \endfoot
    \bottomrule
    \endlastfoot
    \multicolumn{5}{l}{\textit{Text classification}} \\
    \midrule
    News Cat & \citep{McemilgNewsCat}\newline{\scriptsize\href{https://huggingface.co/datasets/mcemilg/news-cat}{mcemilg/news-cat}} & Not declared & Unclear & Used as-is; standardized split \\
    Product Reviews & \citep{FthbrmnbyTurkishProductReviews}\newline{\scriptsize\href{https://huggingface.co/datasets/fthbrmnby/turkish_product_reviews}{fthbrmnby/turkish\_product\_reviews}} & CC~BY-SA~4.0, stated in the card's licensing section and in the provider's GitHub LICENCE file (the card's license tag is \texttt{unknown}) & Yes & Used as-is; standardized split \\
    Bil Tweet News & \citep{Toraman2017EventPrediction,Toraman2021EventRetrieval}\newline{\scriptsize\href{https://huggingface.co/datasets/ctoraman/BilTweetNews-sentiment-analysis}{ctoraman/BilTweetNews-sentiment-analysis}} & CC~BY-NC-SA~4.0 & Non-commercial only & Used as-is; standardized split \\
    Gender Hate Speech Turkish & \citep{Toraman2022HateSpeech,Sahinuc2023GenderBias}\newline{\scriptsize\href{https://huggingface.co/datasets/ctoraman/gender-hate-speech-turkish}{ctoraman/gender-hate-speech-turkish}} & CC~BY-NC-SA~4.0 & Non-commercial only & Used as-is; standardized split \\
    \midrule
    \multicolumn{5}{l}{\textit{Token classification}} \\
    \midrule
    WikiNER & \citep{Altinok2023Turkish}\newline{\scriptsize\href{https://huggingface.co/datasets/turkish-nlp-suite/turkish-wikiNER}{turkish-nlp-suite/turkish-wikiNER}} & CC~BY-SA~4.0; the sentences are re-annotated samples of the Kuzgunlar NER dataset, so share-alike and the upstream terms apply & Yes, share-alike & Used as-is; standardized split \\
    WikiANN-TR & \citep{Rahimi2019WikiANN}\newline{\scriptsize\href{https://huggingface.co/datasets/unimelb-nlp/wikiann}{unimelb-nlp/wikiann}} & License tag \texttt{unknown} on the HF card; text is Wikipedia-derived (CC~BY-SA) & Unclear & Used as-is; standardized split \\
    PosUD-BOUN & \citep{UDTurkishBOUN}\newline{\scriptsize\href{https://github.com/UniversalDependencies/UD_Turkish-BOUN}{UniversalDependencies/UD\_Turkish-BOUN}} & CC~BY-SA~4.0 & Yes & Used as-is; standardized split \\
    PosUD-IMST & \citep{UDTurkishIMST}\newline{\scriptsize\href{https://github.com/UniversalDependencies/UD_Turkish-IMST}{UniversalDependencies/UD\_Turkish-IMST}} & CC~BY-NC-SA~4.0 & Non-commercial only & Used as-is; standardized split \\
    \midrule
    \multicolumn{5}{l}{\textit{Semantic textual similarity}} \\
    \midrule
    SICK-TR & \citep{Dehghan2025Turkish}\newline{\scriptsize\href{https://github.com/SoDehghan/SICK-TR}{SoDehghan/SICK-TR}} & MIT & Yes & Used as-is; standardized split \\
    STSb-TR & \citep{BekenFikri2021Summarization}\newline{\scriptsize\href{https://github.com/verimsu/STSb-TR}{verimsu/STSb-TR}} & CC~BY-SA~4.0 for the similarity scores; underlying source text licensing mixed and not separately cleared & Partial / mixed & Used as-is; standardized split \\
    \midrule
    \multicolumn{5}{l}{\textit{Natural language inference}} \\
    \midrule
    SNLI-TR & \citep{BudurEtAl2020Data}\newline{\scriptsize\href{https://github.com/boun-tabi/NLI-TR}{boun-tabi/NLI-TR}} & CC~BY-SA~4.0, matching the original SNLI license & Yes & Used as-is; standardized split \\
    MultiNLI-TR & \citep{BudurEtAl2020Data}\newline{\scriptsize\href{https://huggingface.co/datasets/boun-tabi/nli_tr}{boun-tabi/nli\_tr}} & Inherits MultiNLI's per-genre terms; HF release tags list cc-by-3.0, cc-by-4.0, cc-by-sa-3.0, MIT and other & Yes, per genre & Used as-is; standardized split \\
    \midrule
    \multicolumn{5}{l}{\textit{Question answering}} \\
    \midrule
    TQuAD & \citep{ErdometoTQuad2}\newline{\scriptsize\href{https://huggingface.co/datasets/erdometo/tquad2}{erdometo/tquad2}} & HF card license \texttt{other} (named \texttt{tquad2}), linking to another copy of the dataset that carries no license statement & Unclear & Used as-is; standardized split \\
    XQuAD & \citep{Artetxe2019CrossLingual}\newline{\scriptsize\href{https://github.com/google-deepmind/xquad}{google-deepmind/xquad}} & CC~BY-SA~4.0 & Yes & Used as-is; standardized split \\
    \midrule
    \multicolumn{5}{l}{\textit{Information retrieval}} \\
    \midrule
    BiText & \citep{Bojar2016WMT16}, redistributed via~\citep{BaysanGungor2025TR-MTEB}\newline{\scriptsize\href{https://huggingface.co/datasets/trmteb/wmt16_en_tr_fine_tuning_dataset}{trmteb/wmt16\_en\_tr\_fine\_tuning\_dataset}} & Not declared on the redistributed HF card; the WMT16 task page states no licence for the news-translation data, and the SETIMES2 portion is distributed by OPUS & Not declared & Human-translated (WMT16 professional translators + SETIMES2 parallel text); split only \\
    MsMarco-TR, Scifact-TR, Fiqa-TR, NFCorpus-TR, Quora-TR & \citep{BaysanGungor2025TR-MTEB,Muennighoff2023MTEB}\newline{\scriptsize\href{https://huggingface.co/datasets/trmteb/msmarco-tr_fine_tuning_dataset}{trmteb/msmarco-tr\_fine\_tuning\_dataset}}\newline{\scriptsize MsMarco-TR: pre-existing translation collected by TR-MTEB, attributed to \citet{Kesgin2023YTU}; the other four translated by TR-MTEB (\texttt{aya-expanse-8b})} & Not declared on any of the five HF cards (\texttt{trmteb/*-tr}); upstream source corpora carry heterogeneous terms not individually re-audited here. TR-MTEB's authors state that datasets were accessed under their original license terms and that derived resources are released for non-commercial research use only & Research use only & Four translated by TR-MTEB (\texttt{aya-expanse-8b}), MsMarco-TR pre-existing; split only \\
    \midrule
    \multicolumn{5}{l}{\textit{Code retrieval (NL components translated by the authors, GPT-4.1)}} \\
    \midrule
    Apps-TR & \citep{Hendrycks2021Measuring}\newline{\scriptsize\href{https://github.com/hendrycks/apps}{hendrycks/apps}} & MIT & Yes & Problem text translated; code unchanged; split \\
    CodeSearchNet-21K-TR & \citep{Husain2019CodeSearchNet}\newline{\scriptsize\href{https://github.com/github/CodeSearchNet}{github/CodeSearchNet}} & MIT & Yes, per repo terms & Subsampled; docstrings translated; split \\
    CosQA-TR & \citep{Huang2021CosQA}\newline{\scriptsize\href{https://github.com/microsoft/CodeXGLUE}{microsoft/CodeXGLUE}} & Computational Use of Data Agreement (C-UDA) 1.0 & Yes, for computational use & Questions translated; code unchanged; split \\
    StackOverflowQA-TR & \citep{Li2025Coir}\newline{\scriptsize\href{https://huggingface.co/datasets/CoIR-Retrieval/stackoverflow-qa}{CoIR-Retrieval/stackoverflow-qa}} & Not declared (no license tag on the HF card); Stack Overflow's own default content license (CC~BY-SA) was not independently re-confirmed for this derivative & Unclear & NL components translated; split \\
    \midrule
    \multicolumn{5}{l}{\textit{Academic understanding}} \\
    \midrule
    PubMedRCT-10K-TR & \citep{Dernoncourt2017Pubmed}\newline{\scriptsize\href{https://huggingface.co/datasets/armanc/pubmed-rct20k}{armanc/pubmed-rct20k}} & No license stated by the dataset authors; NLM provides PubMed data freely but notes that some abstracts may be protected by copyright & Conditional & Stratified subsample (10K); translated; split \\
    SciCite-TR & \citep{Cohan2019Structural}\newline{\scriptsize\href{https://huggingface.co/datasets/allenai/scicite}{allenai/scicite}} & Apache-2.0 for the authors' GitHub repository, which distributes the data; the HF card's license tag is \texttt{unknown} & Unclear & Translated; standardized split \\
    Thesis-Abstract-Class.-11K & \citep{TurkishThesis2025Ertugrul}\newline{\scriptsize\href{https://huggingface.co/datasets/umutertugrul/turkish-academic-theses-dataset}{umutertugrul/turkish-academic-theses-dataset}} & CC~BY~4.0; the source card is gated (access on request, approved automatically) & Yes & Class-balanced subset curated (already Turkish); split \\
\end{longtable}
\endgroup

\subsection{Translation Audit Protocol}
\label{sec:audit-protocol}
For each of the six datasets we translated, 20 items were drawn uniformly at random from the test split with a fixed seed, excluding items whose ratio of Turkish to English length lies more than two standard deviations from the dataset mean; up to three such items per dataset were sampled separately as the outlier stratum reported in Section~\ref{sec:translation-limitations}. For StackOverflowQA-TR, half of the items were questions and half answers.
Each rater received the same items in a different random order, each showing the English source, the Turkish translation and the item's gold information: the label for classification items and the paired document or answer, in English, for retrieval items. Raters were asked not to confer before submitting.
The rubric defined three levels: fully adequate (meaning preserved; fluent, domain-appropriate Turkish); minor flaws (meaning preserved but over-literal or awkward phrasing, inconsistent terminology, or an English term left where a Turkish one exists); and meaning changed (content added or omitted, including numbers, units or qualifiers; negation flipped; a technical term mistranslated so that the content differs). Code, identifiers and library names were never translated and were not to be counted as errors. A separate field recorded whether the gold label or document still held for the Turkish text.
Items from MedNLI-TR were rated under the same protocol but are excluded, as the dataset is no longer part of \tabibench.

\label{lastpage}
\end{document}